\pdfoutput=1

\documentclass[11pt]{article}

\usepackage[final]{acl}
\usepackage{ulem}

\usepackage{times}
\usepackage{latexsym}
\usepackage{subfig}
\usepackage{algorithm}
\usepackage{algorithmic}
\usepackage{colortbl} 
\usepackage{multirow}
\usepackage{booktabs}
\usepackage{array}
\usepackage{amsmath}
\usepackage{amssymb}
\usepackage{setspace}

\usepackage[T1]{fontenc}

\usepackage[utf8]{inputenc}

\usepackage{microtype}

\usepackage{inconsolata}

\usepackage{graphicx}
\usepackage{enumitem}
\usepackage{booktabs}
\usepackage{xcolor}

\definecolor{bluedot}{rgb}{0.39, 0.58, 0.93}
\definecolor{crimson}{rgb}{0.86, 0.08, 0.24}

\title{TokenSelect: Efficient Long-Context Inference and Length Extrapolation for LLMs via Dynamic Token-Level KV Cache Selection}

\author{%
  \textbf{Wei Wu}\textsuperscript{1}\footnotemark[2],
  \textbf{Zhuoshi Pan}\textsuperscript{2}\footnotemark[2],
  \textbf{Kun Fu}\textsuperscript{3},
  \textbf{Chao Wang}\textsuperscript{1},
  \textbf{Liyi Chen}\textsuperscript{4},\\[2mm]
  \textbf{Yunchu Bai}\textsuperscript{1},
  \textbf{Tianfu Wang}\textsuperscript{5},
  \textbf{Zheng Wang}\textsuperscript{3},
  \textbf{Hui Xiong}\textsuperscript{5,6}\footnotemark[1]\\[2mm]
  \textsuperscript{1}University of Science and Technology of China,\;
  \textsuperscript{2}Tsinghua University,\\
  \textsuperscript{3}Alibaba Cloud Computing,\;
  \textsuperscript{4}Xiaohongshu Inc.,\\
  \textsuperscript{5}The Hong Kong University of Science and Technology (Guangzhou),\\
  \textsuperscript{6}The Hong Kong University of Science and Technology\\[2mm]
  \texttt{urara@mail.ustc.edu.cn, xionghui@ust.hk}
}

\begin{document}
\maketitle
\renewcommand{\thefootnote}{\fnsymbol{footnote}}
\footnotetext[1]{Corresponding Author.}
\footnotetext[2]{Equal Contribution.}
\footnotetext[4]{Code: \url{https://github.com/pzs19/TokenSelect}}

\begin{abstract}
Rapid advances in Large Language Models (LLMs) have spurred demand for processing extended context sequences in contemporary applications. However, this progress faces two challenges: performance degradation due to sequence lengths out-of-distribution, and excessively long inference times caused by the quadratic computational complexity of attention. These issues limit LLMs in long-context scenarios.
In this paper, we propose Dynamic Token-Level KV Cache Selection (\textit{TokenSelect}), a training-free method for efficient and accurate long-context inference. \textit{TokenSelect} builds upon the observation of non-contiguous attention sparsity, using QK dot products to measure per-head KV Cache criticality at token-level. By per-head
soft voting mechanism, \textit{TokenSelect} selectively involves a few critical KV cache tokens in attention calculation without sacrificing accuracy. To further accelerate \textit{TokenSelect}, we design the Selection Cache based on observations of consecutive Query similarity and implemented the efficient Paged Dot Product Kernel, significantly reducing the selection overhead.
A comprehensive evaluation of \textit{TokenSelect} demonstrates up to $23.84\times$ speedup in attention computation and up to $2.28\times$ acceleration in end-to-end latency, while providing superior performance compared to state-of-the-art long-context inference methods.
\end{abstract}
\section{Introduction}
\vspace{-4pt}
With the rapid development of large language models (LLMs), the number of parameters is no longer the sole factor significantly affecting model performance. The ability to effectively process longer context information has become one of the key metrics for evaluating LLMs' capabilities. Recent advances—such as cross-document understanding~\cite{longbench}, LLM-powered search systems~\cite{llm-search-sys}, complex reasoning~\cite{o1}, and other cutting-edge LLM developments~\cite{generator,lemma,sepit,chen2024plan}—have all placed higher demands on the long-context capabilities of LLMs.
There are two main difficulties in using pre-trained LLMs for long-context inference. On one hand, LLMs are limited by their context length during pre-training (\textit{e.g.} Llama 3 only has 8192 tokens). Directly inferencing on longer sequences can lead to severe performance degradation due to reasons including sequence lengths out-of-distribution~\cite{streamllm,lm-inf}. On the other hand, even if LLMs possess sufficiently large context lengths, the quadratic computational complexity of attention with respect to sequence length makes the response time for long-context inference unbearable.

\begin{figure}[t]
\centering
\includegraphics[width=1\columnwidth]{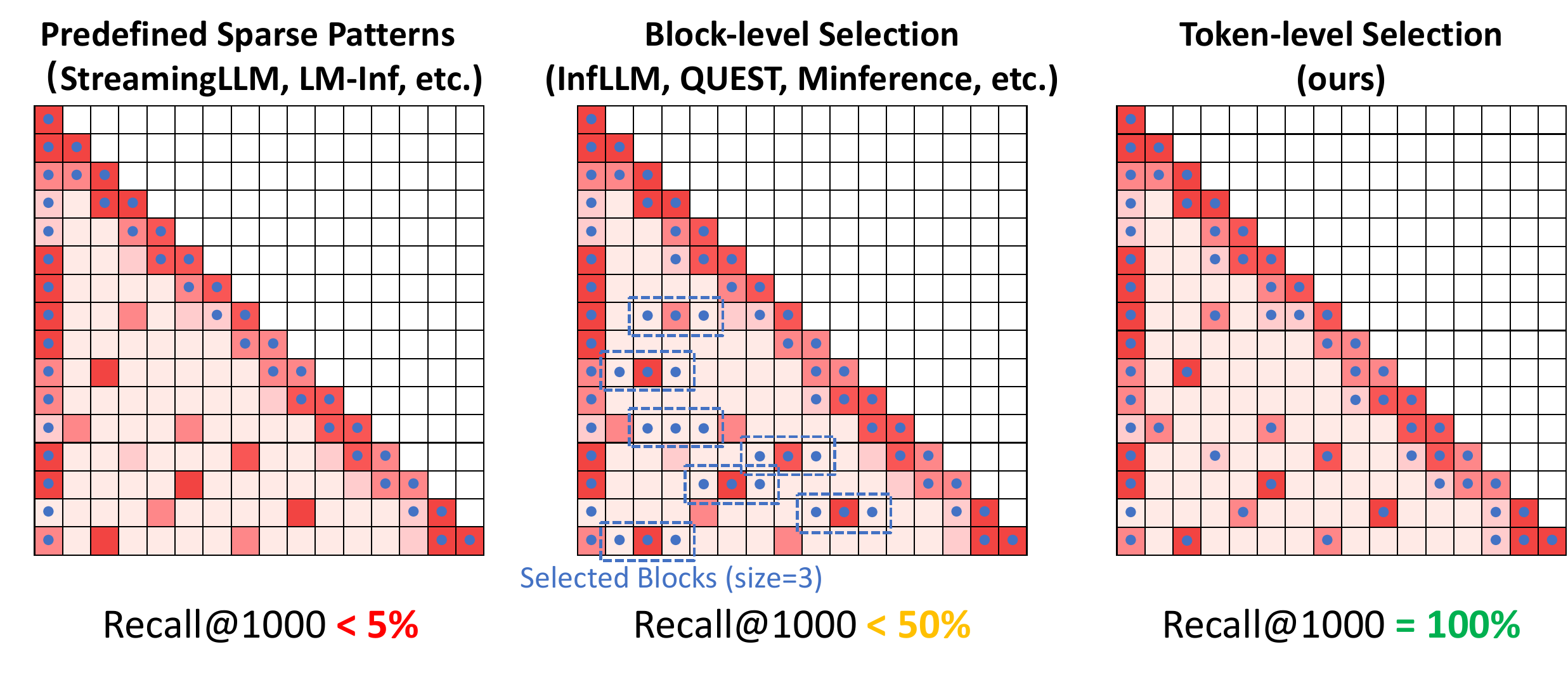}
\vspace{-20pt}
\caption{Distribution of tokens participating in attention computation under different sparsity patterns (\textcolor{bluedot}{blue dots}. \textit{TokenSelect} can more accurately select critical tokens (\textcolor{crimson}{crimson squares}) for attention computation.}
\vspace{-16pt}
\label{fig: intro}
\end{figure}

Previous works have made numerous attempts to address these difficulties. To extend the context length of LLMs, the current common practice is to perform post-training on long texts~\cite{qwen}. However, this approach entails significant computational costs, motivating a training-free and effective method that is computationally efficient. To accelerate long-context inference, many studies focus on the sparsity of attention, attempting to reduce the scale of KV Cache involved in computation. The key to this type of method lies in designing sparse patterns for attention, which can be mainly divided into two categories: one uses predefined sparse patterns~\cite{sliding-windows,big-bird,streamllm,lm-inf}, while the other estimates the potential importance of KV Cache during the inference process~\cite{h2o,snapkv,infinigen,quest,minference,shadowkv}, attempting to select relevant KV Cache tokens into attention calculations. However, the design of these sparse patterns is often heuristically based on historical criticality or coarse-grained criticality estimation of tokens, making it difficult to ensure that the selected tokens are truly critical, thus resulting in sub-optimal performance, as shown in \textit{Fig.} \ref{fig: intro}.

In this paper, we further observe the non-contiguous sparsity of attention, revealing the importance of designing more fine-grained dynamic sparse patterns. To this end, we propose \textit{TokenSelect}, a training-free approach that utilizes token-level selective sparse attention for efficient long-context inference and length extrapolation. 
Specifically, for each Query, \textit{TokenSelect} dynamically calculates token-level per-head criticality for the past KV Cache and selects the $k$ most critical tokens through our head soft vote mechanism, involving them in the attention calculation. This reduces the scale of attention calculation to a constant length familiar to the model, while maintaining almost all of the long-context information, thereby simultaneously addressing the two main difficulties for long-context inference.
To reduce the overhead of token selection, \textit{TokenSelect} manages the KV Cache in token-level pages~\cite{sglang} and design efficient kernel for token selection based on paged KV Cache management through Triton~\cite{triton}. Furthermore, based on our observation of high similarity between consecutive queries, we have designed the Selection Cache, which allows consecutive similar queries to share token selection results, thereby reducing the selection frequency while ensuring its effectiveness. 

We evaluate the performance and efficiency of \textit{TokenSelect} on three representative long-context benchmarks using three open-source LLMs. The experimental results demonstrate that our \textit{TokenSelect} can achieve up to $23.84\times$ speedup in attention computation compared to FlashInfer~\cite{flashinfer}, and up to $2.28\times$ acceleration in end-to-end inference latency compared to state-of-the-art long-context inference method~\cite{infllm}. Simultaneously, it provides superior performance on three long-text benchmarks. In summary, we make the following contributions:

\begin{itemize}[leftmargin=*, labelsep=2mm,itemsep=0mm, topsep=0mm]
    \item An observation on the non-contiguous sparsity of attention that highlights the importance of token-level KV Cache selection.
    \item \textit{TokenSelect}, a training-free method that achieves accurate and efficient long-context inference and length extrapolation, which is compatible with mainstream LLM serving systems.
    \item Comprehensive evaluations of our method, showing up to $23.84\times$ speedup in attention computation and up to $2.28\times$ acceleration in end-to-end latency while exhibiting superior performance.
\end{itemize}

\label{sec:intro}
\section{Related Works}
\label{sec:related_works}
\paragraph{\textbf{Long-context LLMs.}}
Due to computational complexity constraints, current LLMs based on Transformers often utilize limited context lengths during pre-training~\cite{llama2,llama3,mistral,qwen,glm,yi}. To extend their long-context capabilities, existing methods can be broadly categorized into three approaches~\cite{long-context-survey,efficient-inference-survey,length-extrapolation-survey}: 1) Modifying positional encodings: A widely adopted method is positional interpolation~\cite{pi}. ~\citeauthor{pi} first proposed linear scaling of RoPE~\cite{roformer} to map longer positional ranges within the original training window. Subsequent works~\cite{ntk-rope,ntk-rope-dynamic} further improved this method using Neural Tangent Kernel (NTK) theory~\cite{ntk}, achieving longer context windows while maintaining model performance. Methods like YaRN~\cite{yarn} and Giraffe~\cite{giraffe} optimize interpolation effects by adjusting frequency components or introducing temperature parameters. 
2) Long-context post-training: This approach extends the model's context length through additional training steps on longer documents after pre-training~\cite{post-trainin-double,untie-the-knots}. It has been widely adopted by leading LLMs~\cite{gemini,qwen,glm} with the support of sequence parallelism techniques~\cite{megatronlm,deepspeed-ulysses,ringattention}. 
3) Incorporating additional memory modules: Notable examples include Transformer-XL~\cite{transformerxl}, Compressive Transformer~\cite{compressive-transformers}, RMT~\cite{rmt} and Infini-attention~\cite{inf-attention}. Although these methods have expanded the context length of LLMs, long-context inference still faces the challenge of high computational costs.

\paragraph{\textbf{Efficient Long-context Inference.}}
In state-of-the-art LLMs serving systems~\cite{vllm,sglang}, technologies such as Flash Attention~\cite{flashattention2} and Paged Attention~\cite{vllm} have greatly optimized LLMs inference efficiency by improving GPU I/O bottlenecks. However, in long-context inference scenarios, the computational complexity of attention poses new challenges for LLMs inference. Numerous studies focus on the sparsity of attention, selecting partial KV Cache for attention calculations to improve long-context inference efficiency. Sliding window~\cite{sliding-windows,big-bird} is one of the most widely used sparse patterns, reducing complexity to linear by executing attention computations within localized windows. Recent works like StreamingLLM~\cite{streamllm} and LM-infinite~\cite{lm-inf} retain the initial tokens of the sequence in addition to sliding windows, effectively maintaining LLMs' performance when processing long sequences. While these approaches are simple to implement, they cannot retain information from long contexts. Another approach focuses on KV Cache eviction during inference. Methods like H$_2$O~\cite{h2o}, TOVA~\cite{tova} and SnapKV~\cite{snapkv} evaluate token criticality based on historical attention scores, selecting tokens within a limited budget. However, these methods permanently discard parts of the KV Cache, causing information loss from long contexts. To address this, InfLLM~\cite{infllm} introduces Block Memory Units for KV Cache management, retrieving information from long contexts and offloading less-used blocks to CPU. Similarly, QUEST~\cite{quest} proposes query-aware sparsity at page granularity, while MInference~\cite{minference} optimizes long-context inference using three sparse patterns. Apart from considering all attention heads, some other works~\cite{sparq,infinigen} attempt to focus on only a subset of attention heads. While existing methods have shown progress, opportunities for further improvement remain in achieving optimal accuracy and computational efficiency for real-world deployment.

\section{Preliminaries}
\label{sec:preliminaries}
\vspace{-2pt}
As discussed in the \textit{Sec.} \ref{sec:intro}, the high attention sparsity in LLMs suggests sparse attention as a promising solution for long-context inference challenges, which can keep the number of tokens participating in attention computations at a constant scale. Given that predefined sparse patterns are detrimental to performance, we aim to dynamically select crucial tokens at each step during the inference process. Accordingly, based on the overview of LLM inference presented in Appendix~\ref{sec:llm_inference}, we formalize the Selective Sparse Attention Problem as follows.

{\it
\noindent \textbf{Definition 1} (Selective Sparse Attention Problem, informal). For current input of length $C$ ($C=1$ in the decode stage) and KV Cache of length $N$, assuming there are $H$ attention heads with size of $d_h$, let $\mathbf{O}$ be the output of the SDPA:

{\small
\vspace{-4pt}
\begin{equation}
\resizebox{\linewidth}{!}{$
\mathbf{O} = \left[\sigma\left(\frac{\mathbf{Q}^h \cdot {\left[\mathbf{K}^h_\text{\normalfont cache}, \ \mathbf{K}^h_\text{\normalfont current}\right]}^\top}{\sqrt{d}}\right) \cdot [\mathbf{V}^h_\text{\normalfont cache}, \ \mathbf{V}^h_\text{\normalfont current}]\right]_{h=1}^H,
$}
\label{eq:o}
\end{equation}}where $\sigma$ denotes softmax, $\mathbf{Q}^h, \mathbf{K}^h_\text{\normalfont current}, \mathbf{V}^h_\text{\normalfont current} \in \mathbb{R}^{C \times d_h}$ are Query, Key, Value matrices of current input for head $h$ and $\mathbf{K}^h_\text{\normalfont cache}, \mathbf{V}^h_\text{\normalfont cache} \in \mathbb{R}^{N \times d_h}$ represent the KV Cache.
Let $\hat{\mathbf{O}}$ be the output of the Selective Sparse Attention:
{\small
\vspace{-2pt}
\begin{equation}
\resizebox{\linewidth}{!}{$
\hat{\mathbf{O}} = \left[\sigma\left(\frac{\mathbf{Q}^h \cdot {\left[\mathbf{K}^h_\text{\normalfont select}, \ \mathbf{K}^h_\text{\normalfont current}\right]}^\top}{\sqrt{d}}\right) \cdot [\mathbf{V}^h_\text{\normalfont select}, \ \mathbf{V}^h_\text{\normalfont current}]\right]_{h=1}^H,
$}
\label{eq:o_select}
\end{equation}}where $\mathbf{K}^h_\text{\normalfont select},\mathbf{V}^h_\text{\normalfont select} \in \mathbb{R}^{k \times d_h}$ are $k$ selected KV Cache ($k \ll N$). The selection of $\mathbf{K}_\text{\normalfont select}, \mathbf{V}_\text{\normalfont select}$ is performed by selection function $\mathcal{S}$:
{\small
\vspace{-2pt}
\begin{equation}
\begin{aligned}
\mathcal{S}\left(\mathbf{Q},\ \mathbf{K}_\text{\normalfont cache}\right) = \mathcal{I},\; \text{\normalfont where} \; \mathcal{I} \in \mathcal{P}(\{1,\cdots,N\}), \\
\mathbf{K}_{\text{\normalfont select}} = [(\mathbf{K}_{\text{\normalfont cache}})_i]_{i \in \mathcal{I}}, \;
\mathbf{V}_{\text{\normalfont select}} = [(\mathbf{V}_{\text{\normalfont cache}})_i]_{i \in \mathcal{I}},
\end{aligned}
\vspace{-2pt}
\end{equation}}
where $\mathcal{I}$ is the set of selected indices. The objective is to find an appropriate selection function $\mathcal{S}$ that minimizes the difference between the outputs of the SDPA and the selective sparse attention: 

{\small
\vspace{-2pt}
\begin{equation}
\min_{S} \left\| \mathbf{O} - \hat{\mathbf{O}} \right\|_2^2.
\vspace{-2pt}
\end{equation} 
}}

Existing works on long-context inference can be categorized under the Selective Sparse Attention Problem, with variations in the design of the selection function $\mathcal{S}$. Big-Bird and StreamLLM have developed input-independent selection functions $\mathcal{S}()$, while H$_2$O, TOVA and SnapKV propose Query-independent functions $\mathcal{S}(\mathbf{K}_\text{\normalfont cache})$ for improved performance. Current state-of-the-art methods InfLLM, QUEST and MInference utilize Query-aware selection functions $\mathcal{S}(\mathbf{Q}, \mathbf{K}_\text{\normalfont cache})$. However, these approaches typically select at a block-level, which limits their effectiveness.
\begin{figure*}[!t]
\centering
\subfloat[Attention is sparse in token-level.]{
    \label{fig: attention_score} \includegraphics[width=0.62\columnwidth]{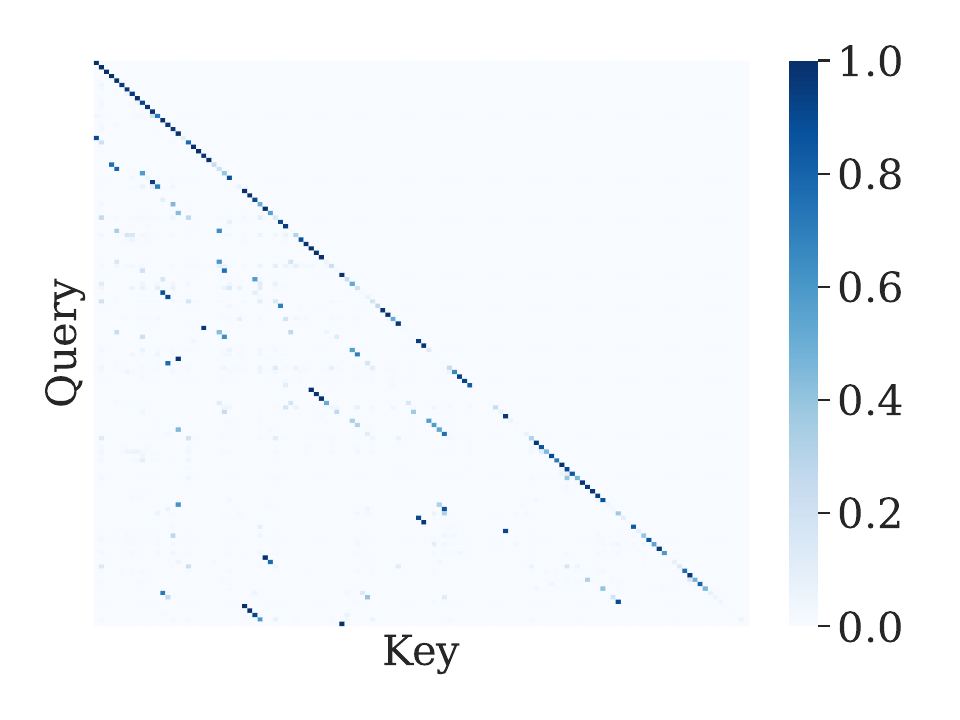}
}
\hspace{0.5em}
\subfloat[Block-level selection is sub-optimal.]{
    \label{fig: recall_wrt_block_size} \includegraphics[width=0.62\columnwidth]{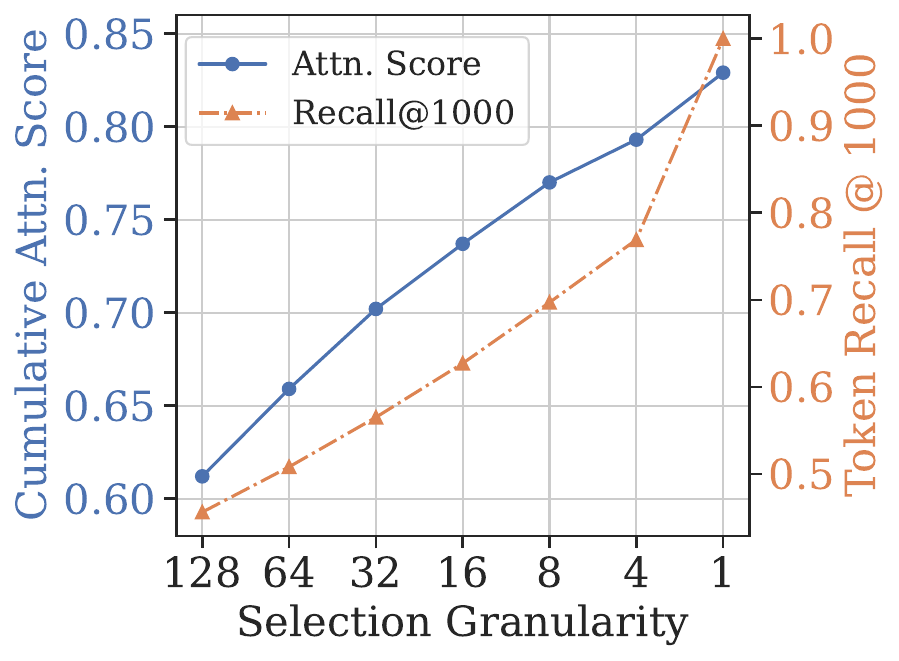}
}
\hspace{0.5em}
\subfloat[Attention logits is head-distinctive.]
{
    \label{fig: attention_logits}
    \includegraphics[width=0.62\columnwidth]{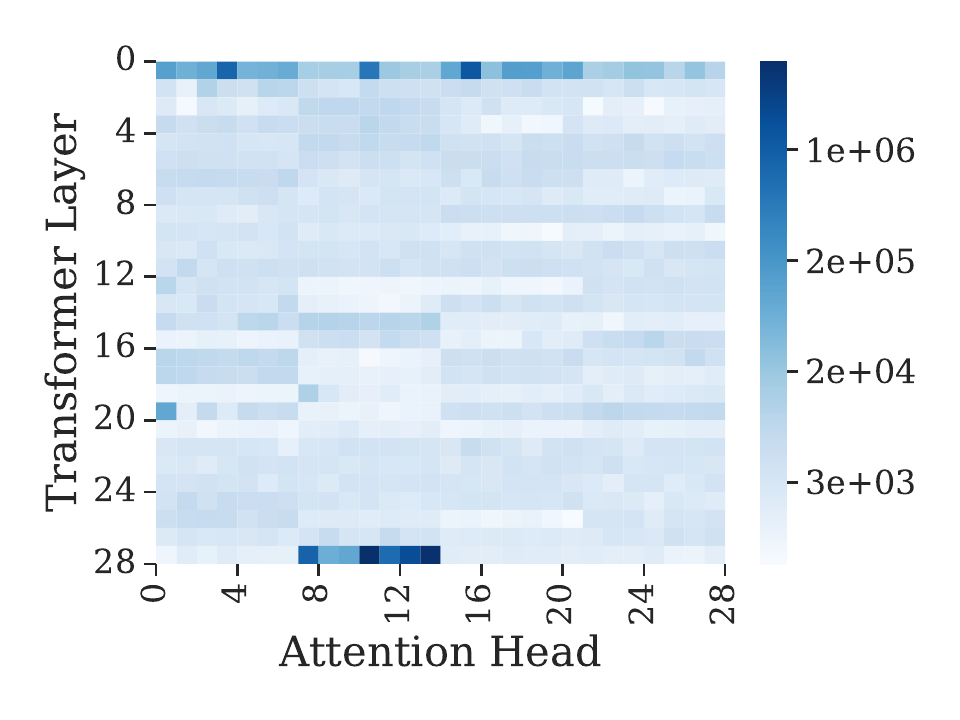}
}
\vspace{-4pt}
  \caption{Motivations for token-level selection. (a) Visualization of attention scores sparsity. (b) Attention scores and critical token recalled by 1K token budget. (c) The $L_1$ norm of attention logits in each attention head.}
  \vspace{-10pt}
  \label{fig: observation1}
\end{figure*}
\begin{figure*}[!t]
\centering

\subfloat[Consecutive queries show consistent similarity patterns across datasets.]
{
    \label{fig: q_sim_across_dataset}
    \includegraphics[width=1.21\columnwidth]{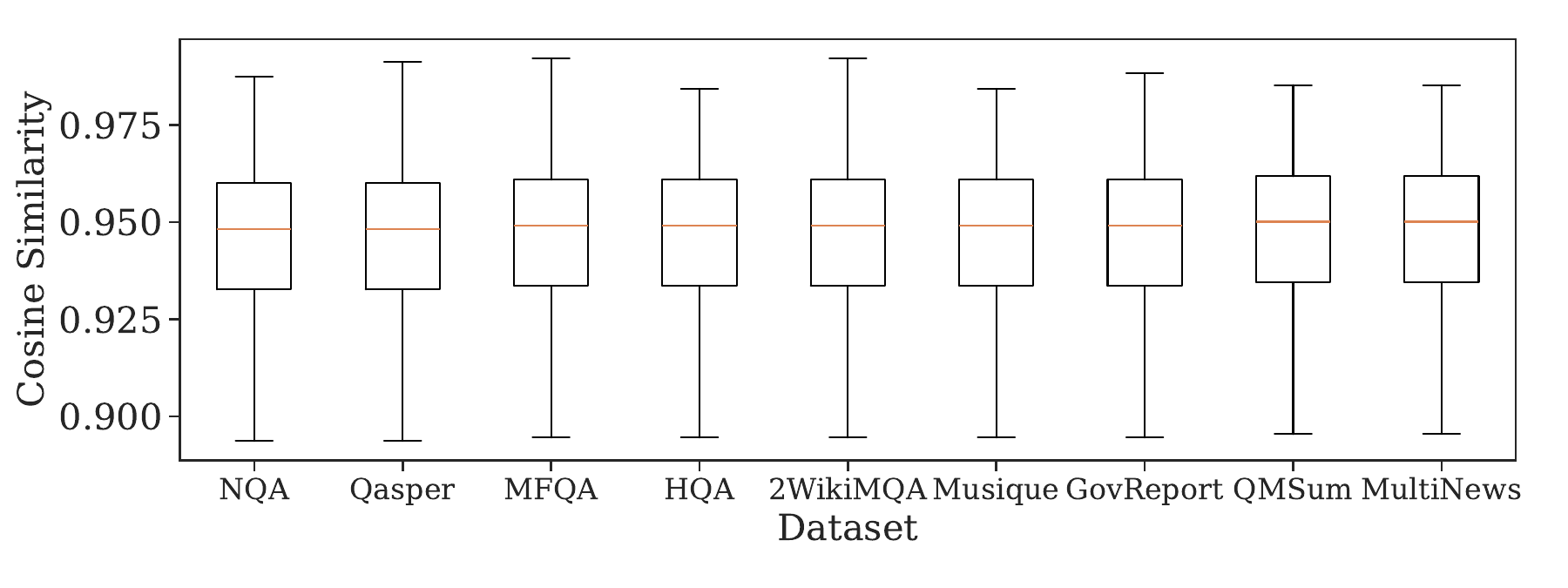}
    \vspace{-10pt}
}
\hspace{0.5em}
\subfloat[Selection overlaps with similar queries.]{
    \label{fig: overlap_rate_wrt_q_sim} \includegraphics[width=0.655\columnwidth]{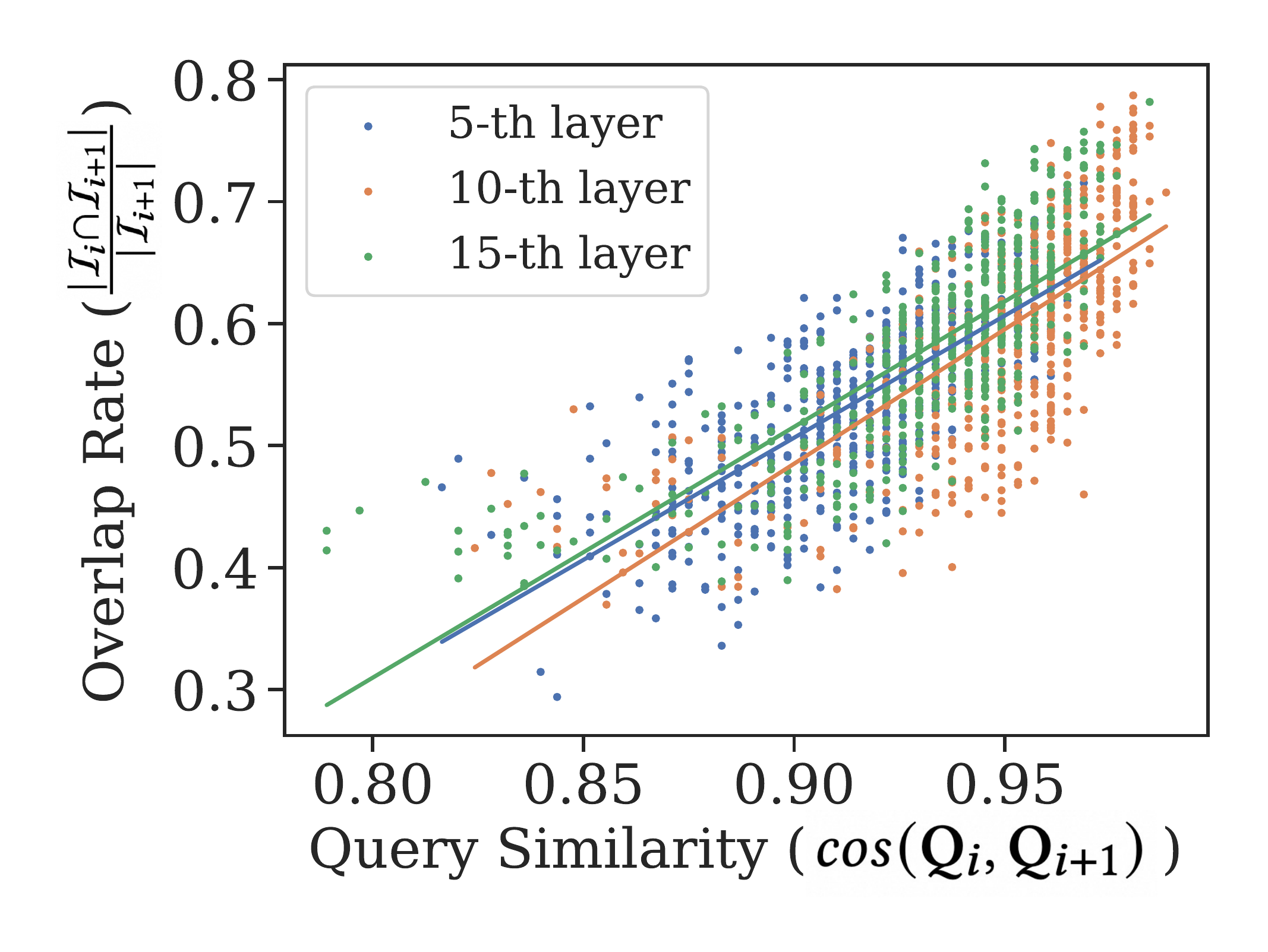}
    \vspace{-10pt}
}
  \vspace{-2pt}
  \caption{Observations on similarity of consecutive queries. (a) Cosine similarity distribution between consecutive queries. (b) The token selection overlap rate ($\frac{|\mathcal{I}_{i} \cap \mathcal{I}_{i+1}|}{|\mathcal{I}_{i+1}|}$) with respect to consecutive Query similarity.}
  \vspace{-4pt}
  \label{fig: observation2}
\end{figure*}
\begin{figure*}[t]
\centering
\includegraphics[width=0.98\textwidth]{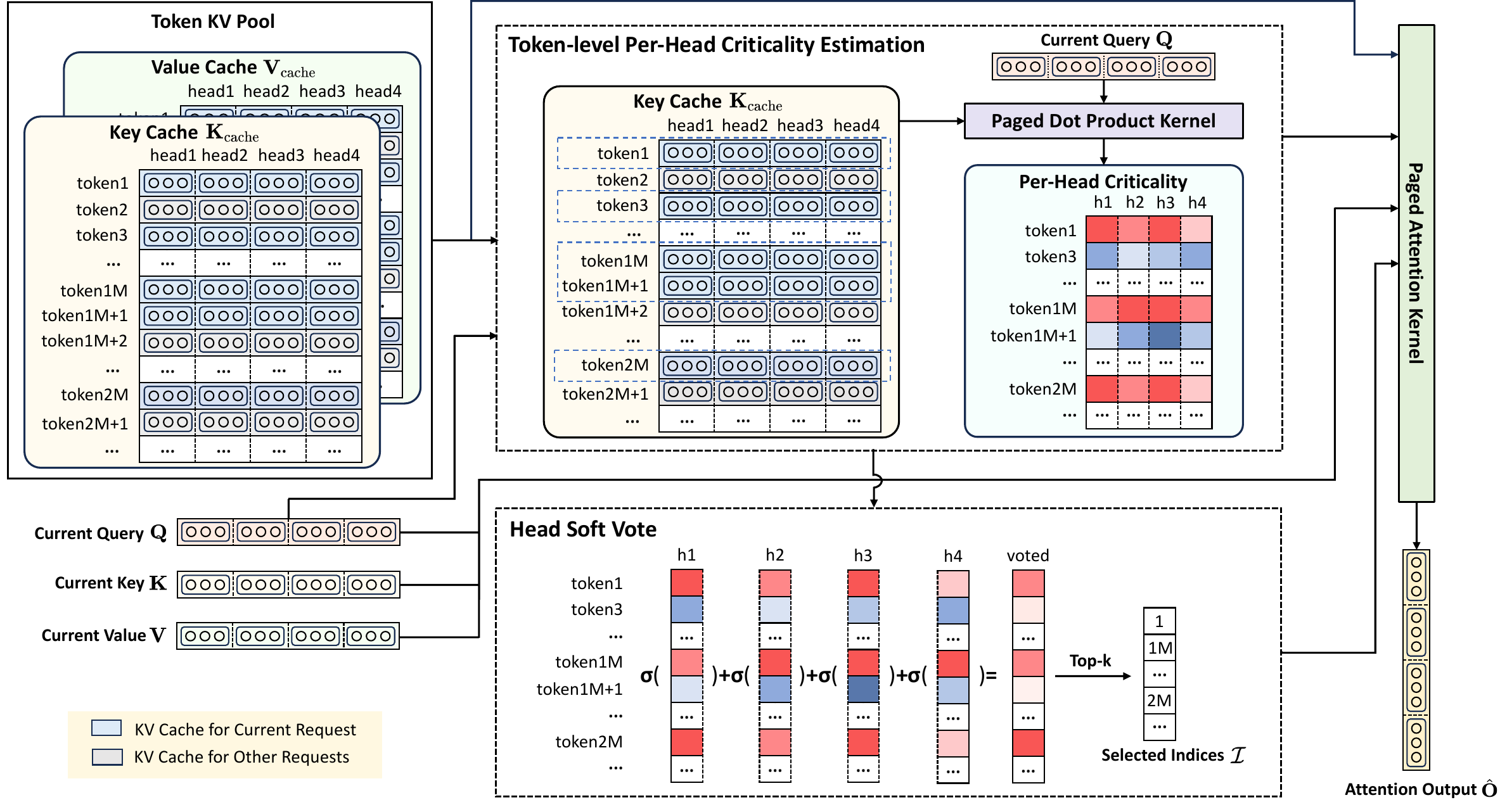}
\vspace{-4pt}
\caption{Execution flow of \textit{TokenSelect}: 
  1) calculate per‐head criticality via Paged Dot Product Kernel; 
  2) perform head soft vote to obtain selected indices; 
  3) execute selective sparse attention via Paged Attention Kernel.}
\vspace{-10pt}
\label{fig: overview}
\end{figure*}

\section{Motivations and Observations}
\paragraph{\textbf{Attention is Sparse, Non-contiguous and Head-Distinctive.}}
\label{sec:attn_obs}

Previous works on long-context inference have demonstrated the sparsity of attention scores in LLMs, particularly when processing long texts. Recent approaches (\textit{e.g.}, InfLLM, QUEST and MInference) partition the KV Cache into non-overlapping blocks, estimating block criticality for sparse attention calculations. These methods assume that critical tokens tend to be contiguous. However, our further observations reveal that this assumption does not always hold true in practice.
As illustrated in \textit{Fig.} \ref{fig: attention_score}, attention scores are sparsely distributed at the token-level. This non-contiguity leads to significant omissions in block-level token selection. \textit{Fig.} \ref{fig: recall_wrt_block_size} demonstrates that finer selection granularity improves recall of critical tokens, motivating us to perform token-level selection. For token-level selection, an intuitive approach would be to directly select the top-$k$ tokens with largest attention logits. However, \textit{Fig.} \ref{fig: attention_logits} reveals considerable disparity in the $L_1$ norm of attention logits across attention heads. As a result, the selection result tends to be dominated by a few heads with disproportionately large attention logits, driving us to design a more robust selection function that maintains the independence of heads.

\paragraph{\textbf{Consecutive Queries are Similar.}}
As sparsity of attention is dynamic~\cite{minference}, token selection should be performed for every Query, which inevitably increases the computational overhead of selective sparse attention. Fortunately, we observe that consecutive Queries exhibit high similarity, as shown in \textit{Fig.} \ref{fig: q_sim_across_dataset}. Intuitively, when two consecutive Queries are highly similar, their dot products with the Keys will also be similar, leading to substantial overlap in the token selection results. Due to space constraints, we provide an informal lemma about this below. The formal version and corresponding proof can be found in the Appendix \ref{sec:app_lemma1}.

{\it \noindent \textbf{Lemma 1} (Informal).
Consider Queries $\mathbf{Q}_{1}, \mathbf{Q}_{2} \in \mathbb{R}^{1\times d}$ that are consecutive and a Key set $\{\mathbf{K}_{i}\}_{i=1}^{N}$. Let $\mathcal{I}_1,$ and $\mathcal{I}_2$ be the sets of indices of the top-$k$ Keys selected by dot product for $\mathbf{Q}_{1},$ and $\mathbf{Q}_{2}$ respectively. If $cos(\mathbf{Q}_{1}, \mathbf{Q}_{2}) > \epsilon$, where $\epsilon$ is a threshold, then $\mathcal{I}_1 = \mathcal{I}_2$.}

\textit{Fig.}~\ref{fig: overlap_rate_wrt_q_sim} illustrates this lemma experimentally. It shows that the overlap rate of token selection tends to increase with Query similarity. This key insight motivates us to reuse selection results for similar queries, improving computational efficiency. 
Moreover, the similarity distribution of consecutive Queries remains consistent across different tasks, as demonstrated in \textit{Fig.}~\ref{fig: q_sim_across_dataset}, allowing us to apply a global similarity threshold across all scenarios. 

\section{Designs of \textit{TokenSelect}}

In this section, we will introduce the design details of \textit{TokenSelect}, primarily encompassing the Selection Function, the Selection Cache, and efficient implementation of \textit{TokenSelect}. The overall workflow of \textit{TokenSelect} is illustrated in \textit{Fig.} \ref{fig: overview}.

\subsection{Selection Function}
The simplest selection function is to determine the criticality of the tokens via the dot product of $\mathbf{Q}$ and $\mathbf{K}_\text{cache}$, then select the top-$k$ as $\mathbf{K}_\text{select},\mathbf{V}_\text{select}$. The selected indices $\mathcal{I}$ are calculated as:

\vspace{-2pt}
{\small
\begin{equation}
\mathcal{I}_\text{topk} = \operatorname{TopK}\left(\mathbf{Q} \cdot {\mathbf{K}_\text{cache}^{h}}^\top\right).
\label{eq:topk}
\vspace{-2pt}
\end{equation}
}

However, as discussed in \textit{Sec.} \ref{sec:attn_obs}, this approach is prone to inaccuracies due to disparities in norm of attention logits between heads. To maintain independence between heads, a better approach is to have each head select the top-$k$ most critical tokens, and then determine the final selection through voting among the heads, where $\mathbb{I}$ is indicator function:

{\small
\vspace{-10pt}
\begin{equation}
\mathcal{I}_{\text {head-vote}}=\operatorname{TopK}\left(\sum_{h=1}^{H} \mathbb{I}\left(i \in \operatorname{TopK}\left(\mathbf{Q}^{h} \cdot{\mathbf{K}_\text{cache}^{h}}^\top\right)\right)\right)
\label{eq:head-vote}
\end{equation}
}
\begin{spacing}{1}
Unfortunately, despite better performance, this method relies on \texttt{scatter\_add} and multiple \texttt{topk} operations, resulting in low efficiency on GPUs. Additionally, the 0/1 voting ignores the relative importance of tokens for each head. Therefore, we propose a head soft vote approach that offers better performance and efficiency. Specifically, we first calculate the per-head criticality, then normalize through softmax, and sum the results for all heads:
\end{spacing}
{\small
\vspace{-12pt}
\begin{equation}
\mathcal{I}_\text{head-soft-vote} = \operatorname{TopK}\left(\sum_{h=1}^{H} \sigma\left(\mathbf{Q}^h \cdot {\mathbf{K}_\text{cache}^{h}}^\top\right)\right).
\label{eq:head-soft-vote}
\vspace{-2pt}
\end{equation} 
}

\subsection{Optimizing Selection Frequency}
\label{sec:selection_frequency}
Although the aforementioned selection function can reduce the complexity of attention from $O(N^2)$ to $O(k^2),k \ll N$, while maintaining performance, the execution time of the selection function itself still affects the latency of inference. To further accelerate long-context inference, based on our observations of the similarity of consecutive Queries, we design optimization strategies for both the prefill stage and the decode stage to reduce the selection frequency while ensuring its effectiveness.

In the prefill stage, $\mathbf{Q}_\text{prefill} \in \mathbb{R}^{n_\text{in} 
\times d}$ is inputed. In long-context scenarios, the number of tokens in the user's input sequence $n_\text{in}$ may reach up to 1M, making it impractical to perform selection for each Query token. Considering the similarity of consecutive Queries, we use chunk-wise token selection, inputting $\frac{1}{c} \sum_{i=1}^{c} (\mathbf{Q}_{C})_i$ into the selection function, where $\mathbf{Q}_C \in \mathbb{R}^{c \times d}$ is the Query chunk and $c$ is the chunk size. This method helps maintain the compute-intensive nature of the prefill stage, preventing it from becoming memory bound.

In the decode stage, due to the auto-regressive characteristic of LLMs, we need to frequently perform selection for $\mathbf{Q}_\text{decode}$, and this process cannot be executed chunk-wise like in the prefill stage. To reduce the frequency of token selection in the decode stage, we propose the Selection Cache. Consecutive similar Queries will hit the cache, thereby directly loading the cached selection results for the previous Query. The Selection Cache allows us to reduce decode latency while maintaining the performance. The formal formulation of the Selection Cache is detailed in Algorithm~\ref{alg:selection_cache}.

\begin{figure}[t]
\centering
\includegraphics[width=1\columnwidth]{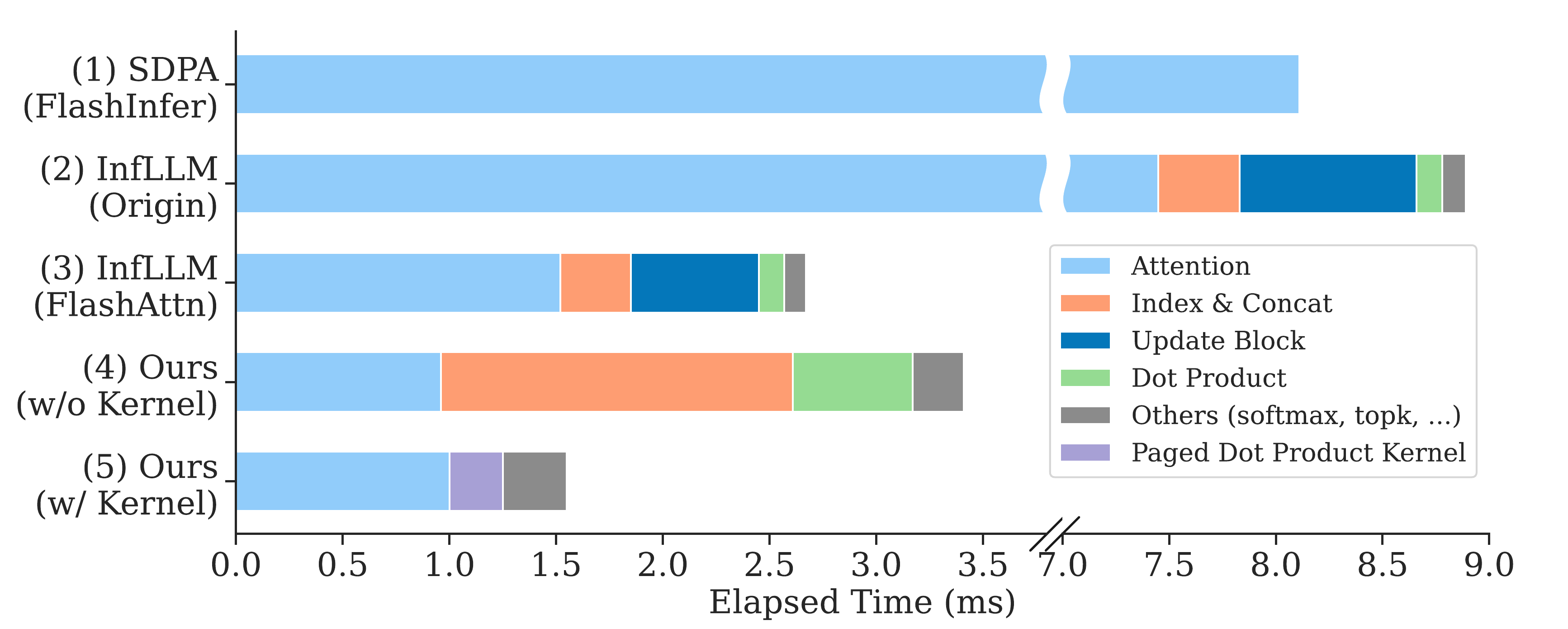}
\vspace{-16pt}
\caption{Time breakdown for single chunk prefill step under different attention implementations (chunk size: 512, KV Cache length: 128K, attended tokens: 4K).}
\vspace{-12pt}
\label{fig: time_break}
\end{figure}
\begin{table*}[t]
\tiny 
\setlength{\arrayrulewidth}{0.3pt}
\setlength{\heavyrulewidth}{0.6pt}
\setlength{\lightrulewidth}{0.3pt}
\renewcommand{\arraystretch}{0.7}
\centering
\resizebox{1.0\textwidth}{!}{
    \begin{tabular}{lcccccccccc}
    \toprule
    Methods & En.Sum & En.QA & En.MC & En.Dia & Code.D & Math.F & R.PK & R.Num & R.KV & Avg. \\
    \midrule
    \textit{Qwen2-7B} & 23.80 & 14.92 & 54.59 & 8.50 & 28.17 & 19.71 & 28.81 & 28.64 & 19.00 & 25.13 \\
    NTK & 18.73 & 15.34 & 41.28 & \bf 7.50 & 24.87 & \bf 27.71 & 99.15 & 97.46 & 59.80 & 43.54 \\
    SelfExtend & 3.76 & 4.44 & 20.09 & 5.00 & 8.12 & 2.29 & 0.00 & 0.00 & 0.00 & 4.86 \\
    StreamingLLM & 19.60 & 13.61 & 48.03 & 3.50 & 27.92 & 19.43 & 5.08 & 5.08 & 2.40 & 16.07 \\
    InfLLM & 19.65 & 15.71 & 46.29 & \bf 7.50 & 27.41 & 24.00 & 70.34 & 72.20 & 5.40 & 32.06 \\
    \rowcolor{cyan!20} \textbf{TokenSelect} & \bf 22.62 & \bf 18.86 & \bf 54.31 & \bf 7.50 & \bf 30.20 &  21.71 & \bf 100.00 & \bf 100.00 & \bf  86.60 & \bf 49.08 \\
    \midrule
    \textit{Llama-3-8B} & 24.70 & 15.50 & 44.10 & 7.50 & 27.92 & 21.70 & 8.50 & 7.80 & 6.20 & 18.21 \\
    NTK & 6.40 & 0.40 & 0.00 & 0.00 & 0.50 & 2.60 & 0.00 & 0.00 & 0.00 & 1.10 \\
    SelfExtend & 14.70 & 8.60 & 19.70 & 0.00 & 0.00 & 22.60 & \bf 100.00 & \bf 100.00 & 0.20 & 29.53 \\
    StreamingLLM & 20.40 & 14.30 & 40.60 & 5.00 & \bf 28.43 & 21.40 & 8.50 & 8.30 & 0.40 & 16.37 \\
    InfLLM & 24.30 & 19.50 & 43.70 & \bf 10.50 & 27.41 & 23.70 & \bf 100.00 & 99.00 & 5.00 & 39.23 \\
    \rowcolor{cyan!20} \textbf{TokenSelect} & \bf 26.99 & \bf 21.32 & \bf 
    45.85 & 8.00 & 27.41 & \bf 28.29 & \bf 100.00 & 97.29 & \bf 48.40 & \bf 43.90 \\
    \midrule
    \textit{Yi-1.5-6B} & 18.78 & 10.48 & 39.74 & 5.00 & 29.95 & 16.00 & 5.08 & 5.08 & 0.00 & 14.45 \\
    NTK & 4.66 & 0.58 & 0.87 & 0.00 & 0.00 & 1.43 & 0.00 & 0.00 & 0.00 & 0.83 \\
    SelfExtend & 5.62 & 1.07 & 1.31 & 0.00 & 0.00 & 1.14 & 0.00 & 0.00 & 0.00 & 1.01 \\
    StreamingLLM & 15.35 & 9.26 & 35.81 & 5.00 & 27.41 & 14.29 & 5.08 & 4.92 & 0.00 & 13.01 \\
    InfLLM & 16.98 & 8.93 & 34.06 & 3.00 & 27.41 & 16.86 & \bf 100.00 & 96.61 & 0.00 & 33.76 \\
    \rowcolor{cyan!20} \textbf{TokenSelect} & \bf 21.13 & \bf 12.32 & \bf 40.61 & \bf 5.50 & \bf 30.71 & \bf 20.86 & \bf 100.00 & \bf 99.83 & 0.00 & \bf 36.77 \\
    \bottomrule
    \end{tabular}
}
\vspace{-2pt}
\caption{Comparison of different methods with different origin models on InfiniteBench.}
\label{tab: infinitebench}
\vspace{-8pt}
\end{table*}

\subsection{Efficient Implementation}
\label{sec:efficient_implementation}

To ready \textit{TokenSelect} for real‐world use, efficient implementation is crucial. We first analyze the time breakdown of representative block-level selective sparse attention method, InfLLM~\cite{infllm}. 
From (1)(2)(3) in \textit{Fig.} \ref{fig: time_break}, we can observe that, despite lowering theoretical complexity, actual runtime depends heavily on implementation. The incompatibility with efficient attention implementations such as Flash Attention has resulted in methods requiring historical attention scores (\textit{e.g.}, H$_2$O, TOVA, SnapKV, InfLLM) impractical in real-world serving.
Analysis of InfLLM’s Flash Attention–compatible version shows that, although block-level criticality estimation aims to cut selection overhead, the dot product isn’t the main bottleneck. Instead, indexing and coalescing selected KV Cache tokens in GPU memory (HBM)—during block updates and KV Cache concatenation—incurs heavy I/O, aggravating LLM inference’s memory-bound limits.
Based on this, we propose that Paged Attention is a more suitable implementation for selective sparse attention. Using paged KV Cache management (with page size=1 for \textit{TokenSelect}), we can reduce the I/O volume for selection results from the scale of all selected KV Caches $O(2kd)$ to the scale of their indices $O(k)$. However, (4) in \textit{Fig.} \ref{fig: time_break} reveals another bottleneck under paged KV Cache management. Since logically contiguous KV Cache is not entirely contiguous in HBM, it also needs to be made contiguous before performing selection operations. To address this issue, we design a Paged Dot Product Kernel using Triton, which significantly improves the overall efficiency of \textit{TokenSelect}. The formal description of this kernel is detailed in Algorithm~\ref{alg:paged_dot}.
\section{Experiments}
In this section, we introduce the experimental setup and evaluate the performance and efficiency of our \textit{TokenSelect} on long-context inference benchmarks.

\subsection{Experimental Settings}
\paragraph{\textbf{Datasets.}} To evaluate \textit{TokenSelect}'s performance on long-context inference, we use three representative datasets: InfiniteBench~\cite{infbench}, RULER~\cite{ruler}, and LongBench~\cite{longbench}. Detailed descriptions and the evaluation metrics used are provided in Appendix \ref{sec:app_metrics}.

\paragraph{\textbf{Baselines.}} 
To conduct a comprehensive evaluation of \textit{TokenSelect}'s performance, we carry out benchmarks on three mainstream open-source LLMs-\texttt{Qwen2-7B-Instruct}~\cite{qwen}, \texttt{Llama-3-8B-Instruct}~\cite{llama3}, and \texttt{Yi-1.5-6B-Chat}~\cite{yi}-comparing against the following state-of-the-art long-context inference methods: NTK-scaled RoPE, SelfExtend, StreamingLLM, InfLLM, \textit{SnapKV}, \textit{InfiniGen}, \textit{QUEST}, \textit{RetrievalAttention} and \textit{MInference}. Detailed descriptions of these methods are provided in Appendix~\ref{sec:des_baseline}. It is worth noting that the methods indicated in \textit{italics} lack length-extrapolation capability; thus, we evaluate them using an alternative approach, applying them to \texttt{Llama-3-8B-Instruct-262k} (long-text post-trained \texttt{Llama-3-8B-Instruct}).

\paragraph{\textbf{Implementation details.}} 
In all experiments in this paper, we employ greedy decoding to ensure the reliability of the results. For our \textit{TokenSelect}, we implement it on SGLang~\cite{sglang}, which is a fast serving framework based on Flasherinfer~\cite{flashinfer}. We implement our method using PyTorch~\cite{torch} and Triton~\cite{triton}. We follow the baseline approach, including 128 initial tokens and $n_\text{local}$ most recent tokens in the attention computation in addition to the $k$ selected tokens. For NTK and SelfExtend, we extend the model's context length to 128K. For StreamLLM, we set $n_\text{local} = \text{4K}$ . For InfLLM, we set $k = \text{4K},n_\text{local} = \text{4K}$. For our \textit{TokenSelect}, we set $k = \text{2K},n_\text{local} = \text{512}$ to demonstrate our token-level KV Cache selection allows us to achieve better performance with a smaller token budget. Due to the need to demonstrate the method under different $n_\text{local}$ and $k$, we denote the specific token budgets in the form of $k+n_\text{local}$ if they differ from the aforementioned settings. For InfiniteBench and LongBench, we set the threshold $\theta$ of the Selection Cache to 0.9. We use NVIDIA A100 to conduct all experiments. When inferencing sequences over 1M tokens, we additionally employee tensor parallelism, which is transparent to our \textit{TokenSelect}.

\begin{table}[!t]
\centering
\setlength{\arrayrulewidth}{1.5pt}
\setlength{\heavyrulewidth}{2pt}
\setlength{\lightrulewidth}{1.5pt}
\huge
\resizebox{\columnwidth}{!}{
    \begin{tabular}{lccccccc}
    \toprule
    Methods & 4K & 8K & 16K & 32K & 64K & 128K & Avg. \\
    \midrule       
    \textit{Qwen2-7B} & 90.74 & 84.03 & 80.87 & 79.44 & 74.37 & 64.13 & 78.93 \\
    StreamingLLM & 94.41 & 54.59 & 33.54 & 22.40 & 15.38 & 10.88 & 38.53 \\
    InfLLM (2K+512) & 52.85 & 36.09 & 29.36 & 23.52 & 18.81 & 18.29 & 29.82 \\
    InfLLM (4K+4K) & 55.22 & 52.10 & 40.53 & 29.77 & 21.56 & 18.64 & 36.30 \\
    \rowcolor{cyan!20} \textbf{Ours (2K+512)} & 94.11 & 81.81 & 68.68 & 60.62 & 51.81 & 42.75 & 66.63 \\
    \rowcolor{cyan!20} \textbf{Ours (4K+4K)} & \bf 94.42 & \bf 90.22 & \bf 82.06 & \bf 70.40 & \bf 59.66 & \bf 54.28 & \bf 75.17 \\
    \midrule
    \textit{Llama-3-8B} & 93.79 & 90.23 & 0.09 & 0.00 & 0.00 & 0.00 & 30.69 \\
    StreamingLLM & 93.68 & 54.48 & 33.77 & 20.35 & 14.88 & 11.47 & 38.11 \\
    InfLLM (2K+512) & 79.79 & 52.43 & 40.12 & 33.60 & 25.68 & 23.39 & 42.50 \\
    InfLLM (4K+4K) & 93.79 & 86.11 & 64.33 & 45.39 & 33.13 & 27.81 & 58.43 \\
    \rowcolor{cyan!20} \textbf{Ours (2K+512)} & 93.73 & 82.92 & \bf 71.92 & \bf 65.38 & \bf 59.35 & 33.39 & \bf 67.78 \\
    \rowcolor{cyan!20} \textbf{Ours (4K+4K)} & \bf 93.88 & \bf 90.29 & 70.13 & 57.72 & 48.36 & \bf 39.38 & 66.63 \\
    \midrule
    \textit{Yi-1.5-6B} & 73.12 & 9.09 & 0.37 & 0.01 & 0.00 & 0.01 & 13.77 \\
    StreamingLLM & 72.10 & 33.03 & 21.69 & 15.39 & 12.58 & 12.61 & 27.90 \\
    InfLLM (2K+512) & 59.66 & 36.77 & 27.41 & 24.49 & 21.49 & 21.17 & 31.83 \\
    InfLLM (4K+4K) & 74.81 & 52.57 & 27.65 & 22.83 & 20.19 & 19.48 & 36.26 \\
    \rowcolor{cyan!20} \textbf{Ours (2K+512)} & \bf 75.93 & \bf 59.55 & \bf 49.69 & \bf 42.36 & \bf 34.68 & \bf 31.36 & \bf 48.93 \\
    \bottomrule
    \end{tabular}
}
\caption{Performance comparison on RULER.}
\vspace{-10pt}
\label{tab: ruler}
\end{table}

\begin{figure*}[t]
\centering
\subfloat[LongBench-GovReport.]{
    \label{fig: govreport} \includegraphics[width=0.45\columnwidth]{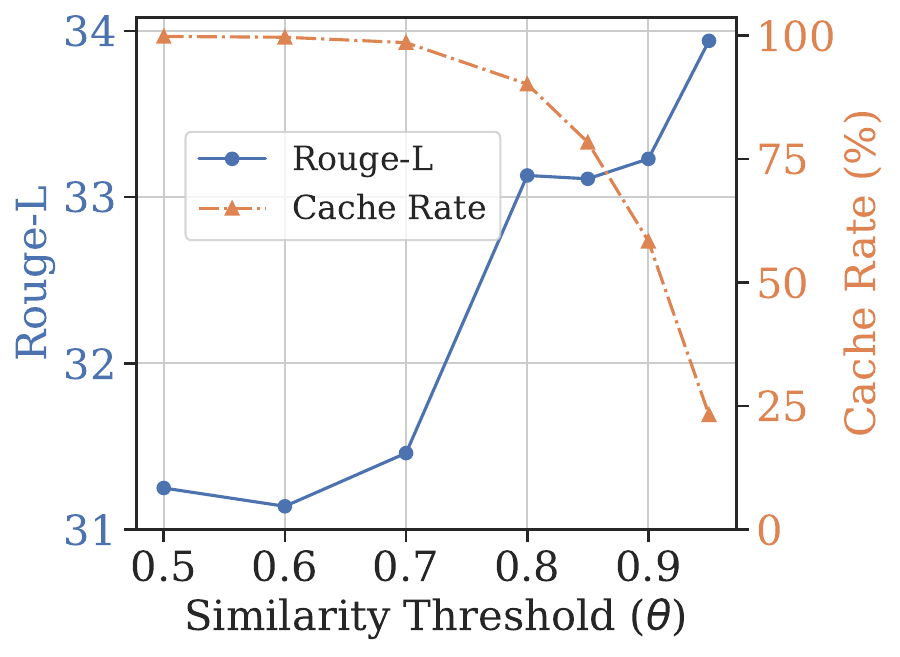}
}
\hspace{0.5em}
\subfloat[InfiniteBench-En.QA.]{
    \label{fig: long_book_qa_eng} \includegraphics[width=0.45\columnwidth]{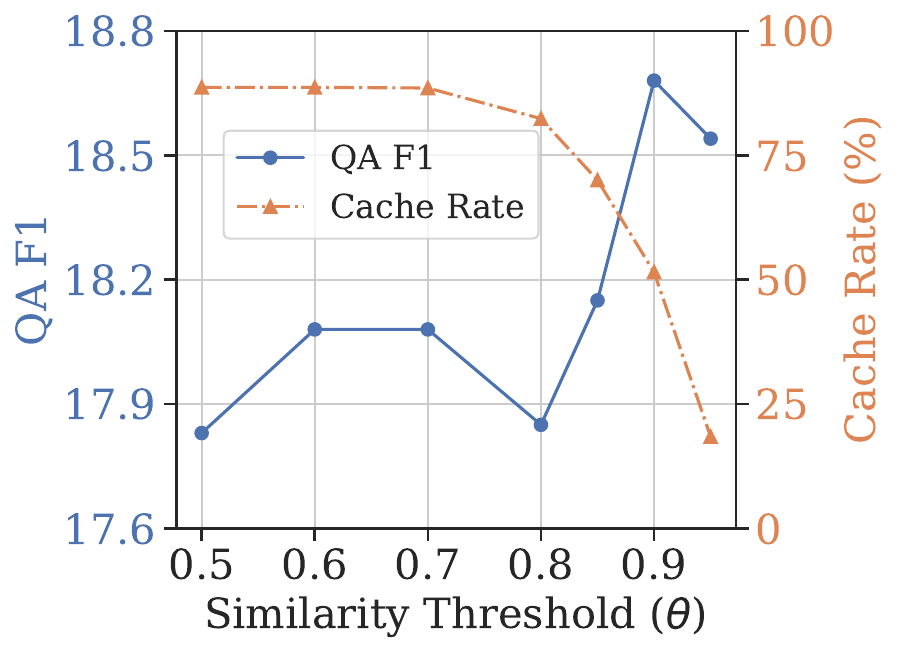}
}
\hspace{0.5em}
\subfloat[InfiniteBench-R.PK.]
{
    \label{fig: passkey}
    \includegraphics[width=0.45\columnwidth]{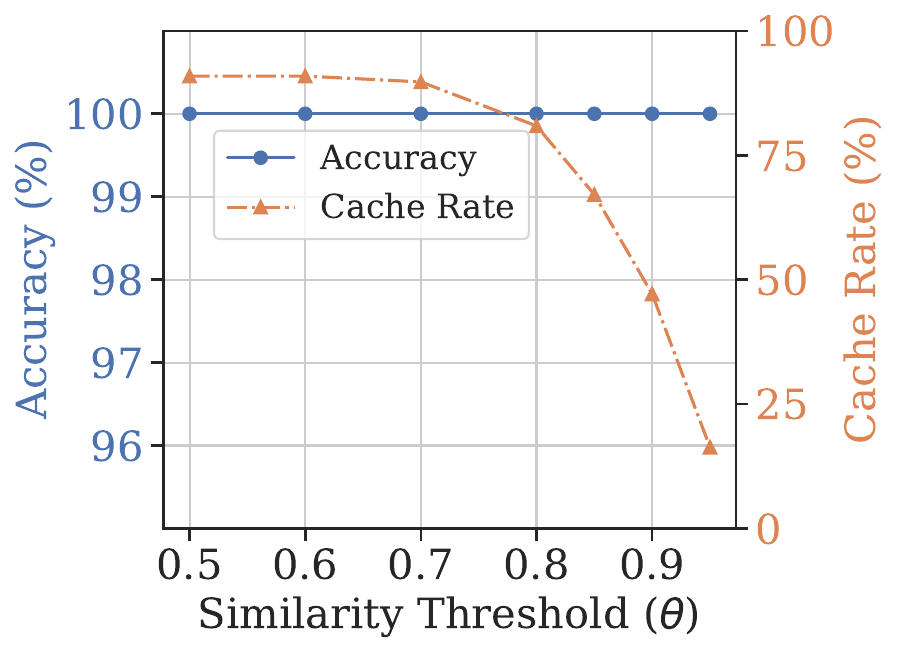}
}
\hspace{0.5em}
\subfloat[InfiniteBench-R.KV.]
{
    \label{fig: kv_retrieval}
    \includegraphics[width=0.45\columnwidth]{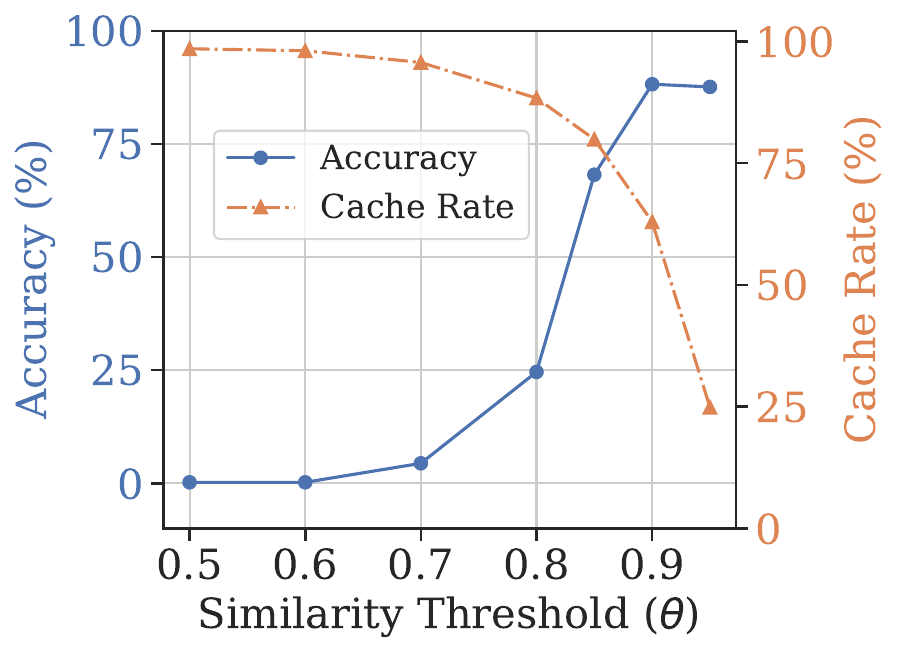}
}
\caption{Performance and Cache Rate with different threshold $\theta$ of the Selection Cache on \texttt{Qwen2-7B-Instruct}.}
\vspace{-4pt}
\label{fig: selection_cache}
\end{figure*}
\subsection{Performance Comparisons}
\paragraph{\textbf{InfiniteBench.}} As shown in Table \ref{tab: infinitebench}, our \textit{TokenSelect} achieves significantly superior overall performance on InfiniteBench compared to all baseline methods, even though \textit{TokenSelect} uses the smallest token budget (<3K). The fact that it significantly outperforms the original models demonstrates \textit{TokenSelect}'s strong length extrapolation capability. We analyze that this is due to our adoption of a fine-grained KV Cache selection strategy, while considering the equal contribution of each head to selection, which ensures that we can select most critical tokens. Observing the performance of other methods, we find that RoPE interpolation methods (NTK, SelfExtend) generally perform poorly unless used on specially trained models such as \texttt{Qwen2-7B-Instruct}. 
The sparse attention method StreamingLLM, based on fixed sparse patterns, can guarantee some of the model's capabilities, but due to discarding a large amount of long-context information, it performs poorly on retrieval-related tasks (R.PK, R.Num, R.KV). The block-level selection method InfLLM can retain more long-context information compared to StreamingLLM. However, due to its sub-optimal block-level selection, it results in lower performance on most tasks compared to \textit{TokenSelect}, even though we set a larger token budget for InfLLM. It is worth noting that \texttt{Yi-1.5-6B} does not perform normally on the R.KV task, as it is unable to correctly recite strings like the UUID.

\paragraph{\textbf{RULER.}} To further demonstrate the capability of \textit{TokenSelect}, we conduct evaluation on the more challenging long-context benchmark RULER. Considering the increased difficulty of RULER and its substantial computational requirements, we include only comparable baseline methods. As shown in Table \ref{tab: ruler}, our \textit{TokenSelect} maintains significantly superior overall performance compared to other long-context inference methods. For all models, \textit{TokenSelect} achieves length extrapolation while preserving the model's original capabilities, benefiting from our efficient utilization of the model's limited context length. Notably, due to the constraints of model's context length, \textit{TokenSelect} experiences performance degradation with larger token budgets (4K+4K) on Llama and Yi. However, its performance with smaller token budgets still significantly surpasses other baseline methods.

\begin{table}[t]
\setlength{\arrayrulewidth}{1.5pt}
\setlength{\heavyrulewidth}{2pt}
\setlength{\lightrulewidth}{1.5pt}
\centering
\Huge
\resizebox{\columnwidth}{!}{
\begin{tabular}{lcccccc}
\toprule
Methods & En.QA & En.MC & Code.D & R.PK & R.Num & R.KV \\
\midrule
\multicolumn{7}{c}{\textit{Llama-3-8B-Instruct-262k}} \\
\midrule
\textit{SDPA (128K)}     & 9.10  & 68.00 & 19.00 & 100.00 & 100.00 & 17.50 \\
\textit{SDPA (262K)}     & 12.40 & 67.30 & 22.10 & 100.00 & 100.00 & 14.40 \\
StreamingLLM (2K+512)    & 6.00  & 66.00 & 18.50 & 5.00   & 5.00   & 1.00  \\
SnapKV (2K+512)          & 11.80 & 67.00 & 18.00 & 100.00 & 100.00 & 0.50  \\
InfLLM (2K+512)          & 7.00  & 37.00 & 20.50 & 100.00 & 100.00 & 0.50  \\
InfiniGen (2K+512)       & 7.30  & 57.50 & 17.50 & 100.00 & 99.50  & 0.00  \\
QUEST (2K+512)           & 8.20  & 67.00 & 18.00 & 100.00 & 100.00 & 0.00  \\
RetrievalAttn. (2K+512)  & 7.50  & 67.00 & 19.00 & 100.00 & 100.00 & 14.00 \\
MInference w/ static     & 8.60  & 43.20 & 20.60 & 92.40  & 96.30  & 0.20  \\
MInference               & 12.90 & 65.90 & 22.30 & 100.00 & 100.00 & 12.80 \\
\rowcolor{cyan!20}
Ours (2k+512)            & 9.70  & \bf 68.00 & 19.00 & \bf 100.00 & \bf 100.00 & 20.60 \\
\midrule
\multicolumn{7}{c}{\textit{Llama-3-8B-Instruct}} \\
\midrule
\rowcolor{cyan!20}
Ours (2k+512)            & \bf 21.32 & 45.85 & \bf 27.41 & \bf 100.00 & 97.29  & \bf 48.40 \\
\bottomrule
\end{tabular}
}
\caption{Performance comparison with methods based-on post-trained
models. Baseline performance is referenced from \citet{minference} and \citet{retrievalattention}.}
\vspace{-8pt}
\label{tab: on_long_model}
\end{table}

\paragraph{\textbf{Comparing to methods based-on post-trained models.}}
In Table \ref{tab: on_long_model}, we present a performance comparison of baseline methods that do not support length extrapolation and must be applied to long-text post-trained models. Our results show that, even compared with models undergoing costly long-text post-training and the methods applied to them, the training-free \textit{TokenSelect} exhibits superior performance on most tasks. These findings further demonstrate the effectiveness of \textit{TokenSelect} in long-context inference and length extrapolation.

\subsection{Ablation Studies}
\paragraph{\textbf{Selection functions $\mathcal{S}$.}}
To compare the performance of different selection functions $\mathcal{S}$ under low token budgets (\textit{i.e.}, token efficiency), we maintain the 2K+512 configuration. From Table \ref{tab: ablation_softvote}, we can observe that our proposed head soft vote mechanism performs significantly better across all tasks. This indicates that using the head soft vote mechanism to balance each head's contribution to token selection results can help us avoid the domination of selection by few heads with large attention logits.
\begin{table}[!t]
\centering
\resizebox{\columnwidth}{!}{
    \begin{tabular}{lccccccc}
    \toprule
     $\mathcal{S}$ & En.QA & En.MC & Code.D & R.PK & R.Num & R.KV  \\
    \midrule
    $\mathcal{I}_{\text{topk}}$ & 15.15 & 45.85 & 28.43 & \bf 100.00 & 98.47 & 16.60 \\
    $\mathcal{I}_{\text{head-vote}}$ & 17.01 & 45.85 & 28.68 & \bf 100.00 & \bf 100.00 & 22.40 \\
    $\mathcal{I}_{\text{head-soft-vote}}$ & \textbf{18.86} & \textbf{54.31} & \textbf{30.20} & \textbf{100.00} & \textbf{100.00} & \textbf{86.60} \\
    \bottomrule
    \end{tabular}
}
\vspace{-4pt}
\caption{Ablation study of the Selection Function $\mathcal{S}$ on InfiniteBench using \texttt{Qwen2-7B-Instruct}.}
\vspace{-8pt}

\label{tab: ablation_softvote}
\end{table}
\begin{table}[t]
\centering
\small
\resizebox{\columnwidth}{!}{
\begin{tabular}{lcccccc}
\toprule
$k$ & En.Sum & En.QA & En.Mc & Math.F & R.Num & R.KV \\
\midrule
128 & 21.23 & 10.46 & 41.48 & 18.00 & 100.00  & 13.40 \\
256 & 22.01 & 11.66 & 41.92 & 19.71 & 100.00 & 20.00 \\
512 & 21.60 & 13.31 & 40.17 & 21.71 & 100.00 & 45.60 \\
1K & 21.35 & 15.13 & 44.10 & 21.71 & 100.00 & 73.00 \\
2K & 22.62 & 18.86 & 54.31 & 21.71 & 100.00 & 86.60 \\
4K & 24.09 & 21.11 & 51.53 & 21.71 & 100.00 & \bf 88.00 \\
8K & 25.32 & 22.93 & 58.52 & 23.71 & 100.00 & 85.40 \\
16K & \bf 26.54 & \bf 23.04 & \bf 62.88 & \bf 28.16 & \textbf{100.00} & 72.00 \\
\bottomrule
\end{tabular}
}
\caption{Performance vs. Number of selected tokens $k$ on InfiniteBench using \texttt{Qwen2-7B-Instruct}.}
\vspace{-28pt}
\label{tab: ablation_k}
\end{table}
\begin{figure*}[t]
\centering
\includegraphics[width=1.2\columnwidth]{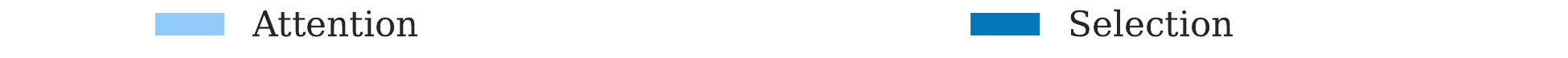}\\
\vspace{-5mm}
\subfloat[KV Cache length: 128K.]{
    \label{fig: 32k} \includegraphics[width=0.6\columnwidth]{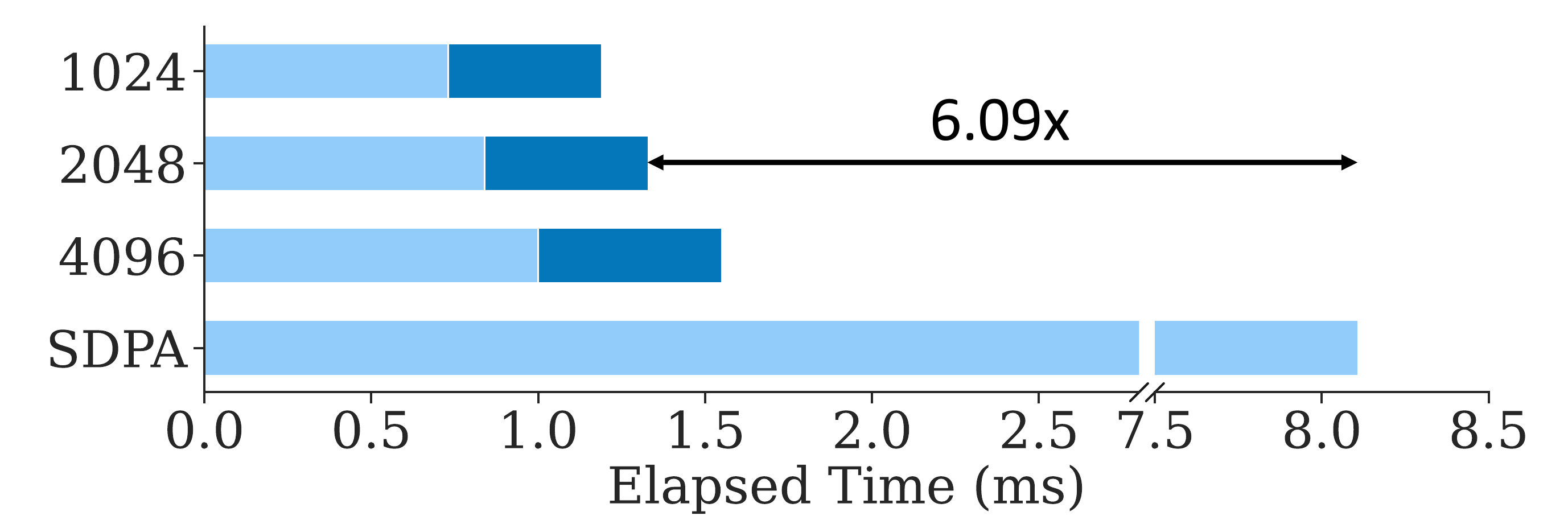}
}
\hspace{0.5em}
\subfloat[KV Cache length: 512K.]{
    \label{fig: 64k} \includegraphics[width=0.6\columnwidth]{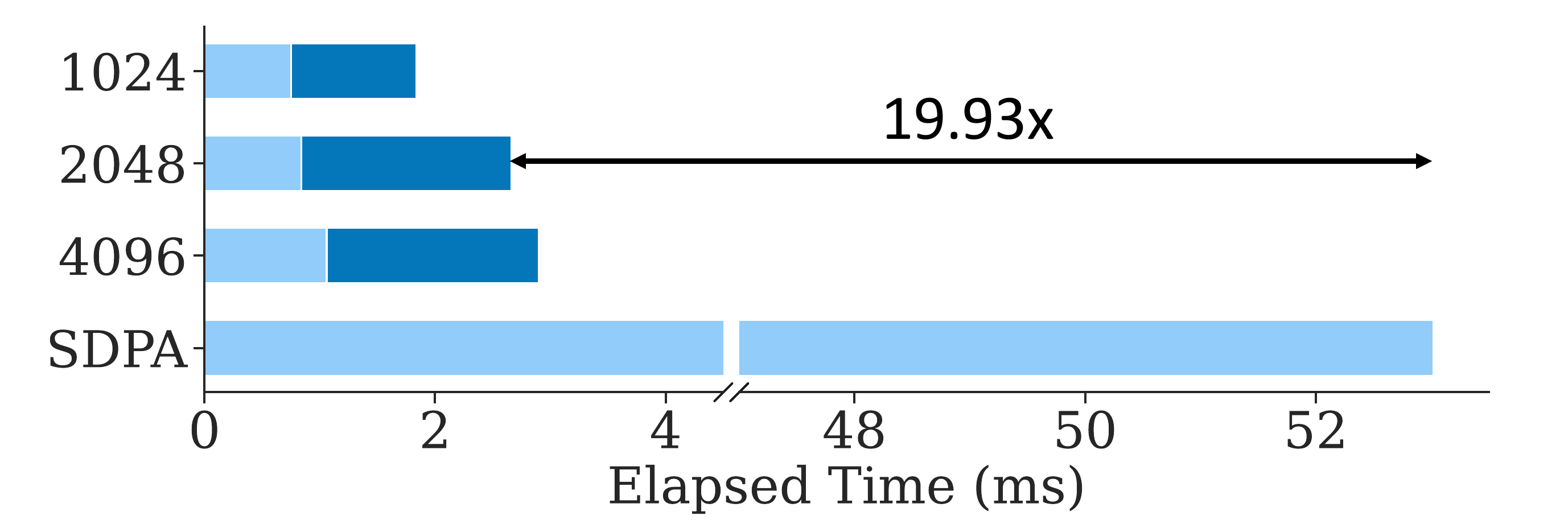}
}
\hspace{0.5em}
\subfloat[KV Cache length: 1M.]
{
    \label{fig: 128k}
    \includegraphics[width=0.6\columnwidth]{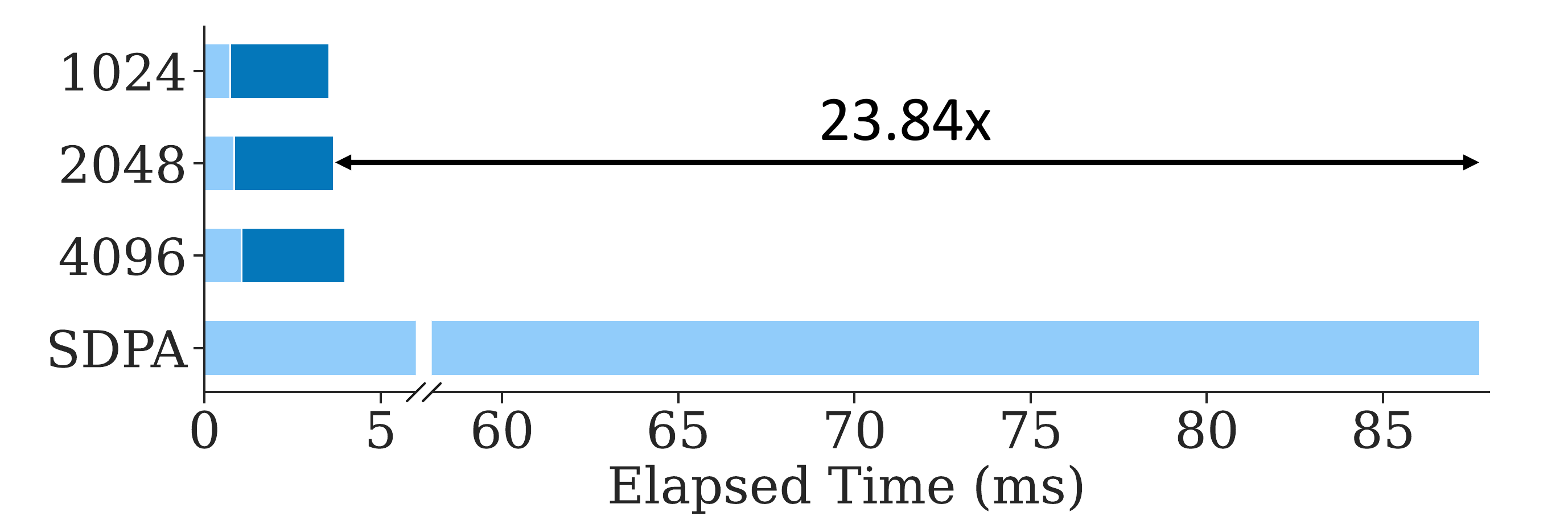}
}
 \vspace{-4pt}
  \caption{Computation time v.s. KV Cache lengths for single chunk prefill step using \texttt{Qwen2-7B-Instruct}. The vertical axis represents the number of attended tokens. SDPA denotes full attention by Flashinfer (chunk size: 512).}
  \vspace{-8pt}
  \label{fig: breakdown_all}
\end{figure*}

\paragraph{\textbf{Similarity threshold of the Selection Cache $\theta$.}} \textit{Fig.} \ref{fig: selection_cache} shows that the Selection Cache hit rate increases significantly as the similarity threshold $\theta$ decreases, converging around $\theta=0.5$. This suggests potential for further acceleration of \textit{TokenSelect}'s decode stage by reducing $\theta$. Performance sensitivity to $\theta$ varies across tasks. While most tasks exhibit slight performance degradation with decreasing $\theta$, and R.PK in InfiniteBench shows no degradation, more challenging retrieval tasks like R.KV demonstrate significant performance deterioration. This indicates higher dynamicity requirements for token selection in these tasks. Owing to the limited generation lengths in current long-context inference benchmarks, we cannot yet precisely quantify the end-to-end speedup provided by the Selection Cache. Nonetheless, for a 7B-parameter model operating on 128K-token sequences, each cache hit reduces per-step latency by approximately 0.5 ms. For more detailed performance comparisons under different $\theta$, see Table \ref{tab: performance_theta} of Appendix \ref{sec:detail_theta}.

\paragraph{\textbf{Number of selected tokens $k$.}}
As shown in Table \ref{tab: ablation_k}, we fix $n_\text{local}$ to a small value ($512$) to compare the performance when selecting different numbers of tokens. First, we observe that even selecting a very small number of tokens (\textit{e.g.}, 128, 256), our \textit{TokenSelect} still demonstrates very comparable performance. Then, as $k$ increases, the effectiveness of \textit{TokenSelect} further improves, indicating that more moderately critical tokens also contribute to the retention of long-context information. Finally, we find that when $k$ is set to larger values (\textit{e.g.}, 16K), our \textit{TokenSelect} shows significant improvements in most tasks, further advancing the performance landscape of long-context inference methods.

\begin{figure}[t]
\centering
\includegraphics[width=0.9\columnwidth]{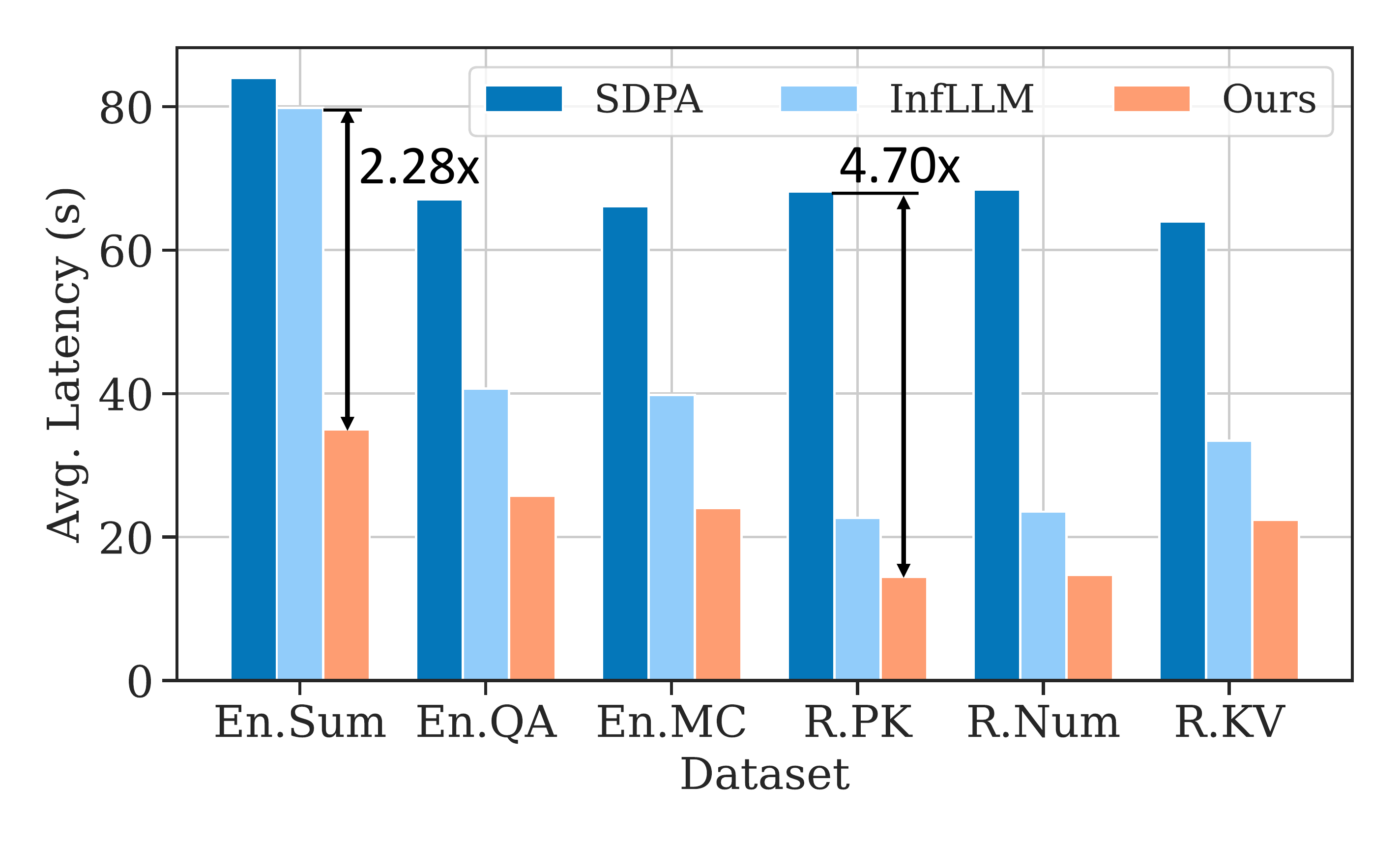}
\vspace{-10pt}
\caption{End to end latency per sample with different methods on InfiniteBench using \texttt{Qwen2-7B-Instruct}.}
\vspace{-8pt}
\label{fig: end2end_latency}
\end{figure}

\subsection{Efficiency Comparisons}
\paragraph{\textbf{Efficiency of selective sparse attention.}} \textit{Fig.} \ref{fig: breakdown_all} demonstrates the significant acceleration of attention computation achieved by \textit{TokenSelect} during long-context inference. With a KV Cache length of 1M, \textit{TokenSelect} can provide up to $23.84\times$ speedup compared to FlashInfer, which is the inference kernel library we based on. This substantial improvement is attributed to our efficient kernel design.

\paragraph{\textbf{End-to-end efficiency.}} \textit{Fig.} \ref{fig: end2end_latency} compares the end-to-end latency of \textit{TokenSelect}, InfLLM, and SDPA across various tasks. \textit{TokenSelect} significantly accelerates long-context inference in real-world scenarios, achieving a maximum speedup of $4.70\times$ over SDPA and $2.28\times$ over the state-of-the-art long-context inference method while also delivering superior overall performance.

\section{Conclusion}
In this paper, we introduces \textit{TokenSelect}, a training-free approach for efficient long-context inference and length extrapolation. \textit{TokenSelect} addresses the two major challenges faced by LLMs in processing long texts: the context length limitation from pre-training and the computational complexity of attention. This is achieved through a novel token-level selective sparse attention mechanism. Experimental results demonstrate that \textit{TokenSelect} can achieve up to $23.84\times$ speedup in attention computation and up to $2.28\times$ acceleration in end-to-end inference latency, while exhibiting superior performance across multiple long-context benchmarks.

\section{Limitations}
Our approach has inherent limitations that present opportunities for future work. A primary limitation of our method is that its training-free design—a significant advantage—acts as a double-edged sword, as its absolute performance is inherently tied to the quality of the underlying LLMs. Although our experiments demonstrate robustness of \textit{TokenSelect} across various LLMs, some inherent shortcomings—such as the misrecognition of UUID strings by \texttt{Yi-1.5-6B-Chat}—indicate that certain issues may still require training to resolve. Moreover, while our method currently achieves state-of-the-art performance in long-context inference, recent long-text post-training techniques in the LLM community have shown impressive performance; notably, our \textit{TokenSelect} is orthogonal to these approaches and can be employed during inference to trade a slight performance drop for significant efficiency gains. Finally, although our method achieves state-of-the-art efficiency improvements in long-context inference, the task remains inherently resource-intensive. For instance, even with a 8B-parameter model, complex benchmarks (e.g., RULER) can require approximately 8$\times$A100 GPUs for nearly one day of runtime, and the computational cost is expected to increase substantially for larger models. We hope that our work, together with the community's advances in model design, algorithm development, and infrastructure optimization, will help pave the way for further mitigating these computational challenges.

\section*{Acknowledgment}
This work was supported in part by the National Key R\&D Program of China (Grant No.2023YFF0725001), in part by the National Natural Science Foundation of China (Grant No.92370204), in part by the Guangdong Basic and Applied Basic Research Foundation (Grant No.2023B1515120057), in part by the Education Bureau of Guangzhou Municipality.

\bibliography{custom}
\newpage
\appendix
\section{Formal Description of Algorithms}
\label{sec:algorithm}
In \textit{Sec.}~\ref{sec:selection_frequency}, we propose the Selection Cache, which shares selection results among similar Queries to reduce selection frequency without sacrificing performance. Formally, it is defined as follows:
\begin{algorithm}[h]
\small
\caption{Selection Cache Algorithm}
\label{alg:selection_cache}
\begin{algorithmic}[1]
\REQUIRE 
$\mathbf{Q}\in\mathbb{R}^{H\times D}$: current query vectors\\
\quad\quad\; $k\in\mathbb{N}$: number of tokens to select\\
\quad\quad\; $\mathbf{C}_Q\in\mathbb{R}^{H\times D}$: cached query vector\\
\quad\quad\; $\mathbf{C}_{\mathcal{I}}\in\{0,\dots,N-1\}^{k}$: cached indices\\
\quad\quad\; $\theta\in[0,1]$: cosine-similarity threshold\\
\quad\quad\; $\mathcal{S}$: selection function (Eq.~\ref{eq:head-soft-vote})\\
\quad\quad\; $f\in\{\text{True},\text{False}\}$: first-query flag (default $\text{True}$)
\ENSURE
\;$\mathcal{I}\in\{0,\dots,N-1\}^{k}$: indices of $k$ selected tokens

\IF{$\mathbf{f}$ \textbf{or} $\text{cos}(\mathbf{Q}, \mathbf{C}_Q) < \theta$}
    \STATE $\mathcal{I} \gets \mathcal{S}(\mathbf{Q}, k)$
    \STATE $\mathbf{C}_\mathcal{I} \gets \mathcal{I}$
    \STATE $\mathbf{C}_Q \gets \mathbf{Q}$
    \STATE $\mathbf{f} \gets \text{False}$
\ELSE
    \STATE $\mathcal{I} \gets C_\mathcal{I}$
\ENDIF

\RETURN $\mathcal{I}$
\end{algorithmic}
\end{algorithm}

In \textit{Sec.}~\ref{sec:efficient_implementation}, we propose the Paged Dot Product Kernel to efficiently perform token-level per-head criticality estimation under the paged KV‐cache management by significantly reducing I/O between HBM and SRAM. Formally, it is defined as follows:
\begin{algorithm}[h]
\small
\caption{Paged Dot Product Kernel}
\label{alg:paged_dot}
\begin{algorithmic}[1]
\REQUIRE
$\mathbf{Q}\in\mathbb{R}^{H\times D}$: current query vectors\\
\quad\quad\; $\mathbf{K}\in\mathbb{R}^{N_\text{kv}\times H_\text{kv}\times D}$: key cache pool\\
\quad\quad\; $\mathbf{I}\in\{0,\dots,N_\text{kv}-1\}^{T}$: indices of relevant tokens\\
\quad\quad\; $H$: number of attention heads\\
\quad\quad\; $H_\text{kv}$: number of KV heads $(H \bmod H_\text{kv} = 0)$\\
\quad\quad\; $D$: head dimension\\
\quad\quad\; $T$: number of relevant tokens $(|\mathbf{I}|=T)$\\
\quad\quad\; $B$: CUDA block size
\ENSURE
$\mathbf{S}\in\mathbb{R}^{H\times T}$: dot product scores
\STATE $N \gets \bigl\lceil T / B \bigr\rceil$ 
\FORALL{$h = 0,\dots,H-1$ \textbf{in parallel}}
    \STATE $q \gets \mathbf{Q}[h,:]$ \textcolor{cyan!}{\COMMENT{to SRAM}}
    \STATE $h_\text{kv} \gets h \bmod H_\text{kv}$ 
    \FORALL{$b = 0,\dots,N-1$ \textbf{in parallel}}
        \STATE $t_0 \gets b \times B$
        \STATE $L \gets \min\!\bigl(B,\;T - t_0\bigr)$ 
        \FOR{$j = 0,\dots,L-1$}
            \STATE $idx \gets \mathbf{I}[\,t_0 + j\,]$ \textcolor{cyan!}{\COMMENT{to SRAM}}
            \STATE $k \gets \mathbf{K}[\,idx,\,h_\text{kv},:\,]$ \textcolor{cyan!}{\COMMENT{to SRAM}}
            \STATE $s \gets \langle q,\,k\rangle$ \textcolor{cyan!}{\COMMENT{in SRAM}}
            \STATE $\mathbf{S}[\,h,\,t_0 + j\,] \gets s$ \textcolor{cyan!}{\COMMENT{to HBM}}
        \ENDFOR
    \ENDFOR
\ENDFOR
\RETURN $\mathbf{S}$
\end{algorithmic}
\end{algorithm}

\section{Scalability of \textit{TokenSelect}}
\subsection{Scaling Beyond 1 Million Context Length}
\begin{figure}[h]
\centering
\includegraphics[width=1\columnwidth]{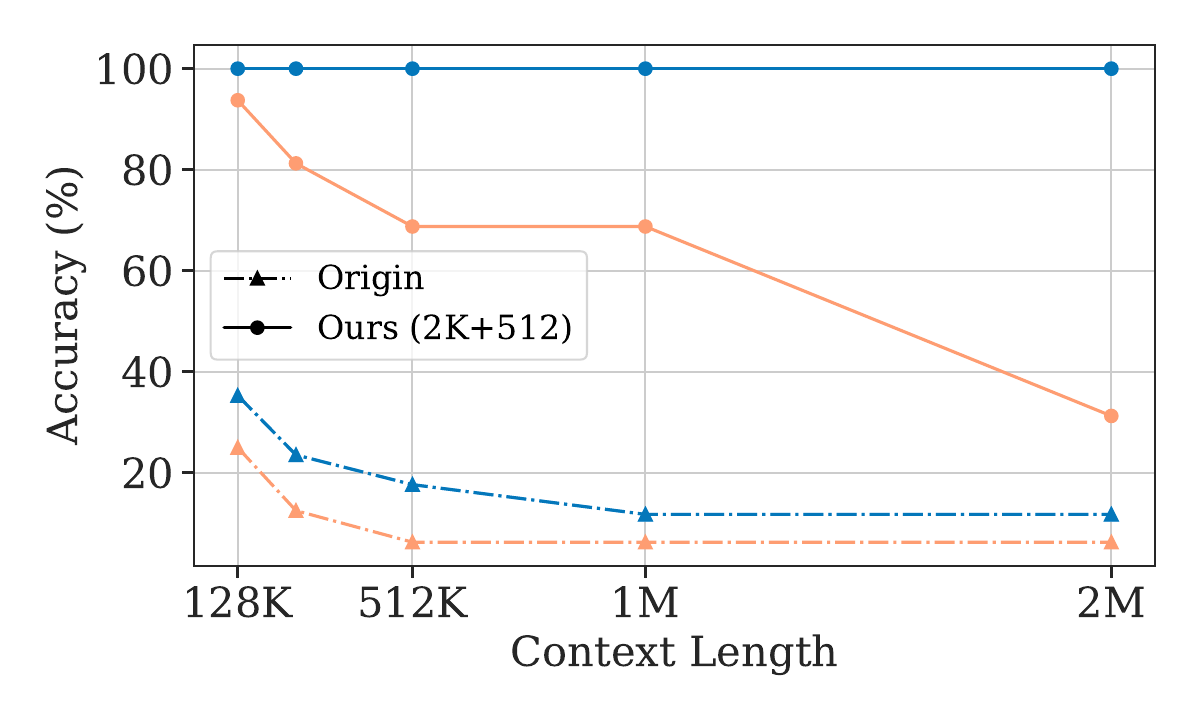}
\caption{Performance comparison on extended \textcolor[RGB]{53,118,181}{R.PK} and \textcolor[RGB]{240,162,124}{R.KV} using \texttt{Qwen2-7B-Instruct}.}
\label{fig: scaling2M}
\end{figure}

To further explore \textit{TokenSelect}'s performance in extreme long-context scenarios, we design an extended benchmark with different text lengths following  InfiniteBench. As illustrated in the \textit{Fig.} \ref{fig: scaling2M}, our \textit{TokenSelect} demonstrates the ability to accurately capture critical information with a small token budget in contexts up to 2M tokens, underscoring its potential in more application scenarios.

\subsection{Scaling to 72 Billion Parameters}
To demonstrate the scalability of our approach to larger models, we conducted additional experiments using \texttt{Qwen2-72B-Instruct}. The results, presented in Table~\ref{tab: longbench_72b}, show that our method outperforms NTK-Aware Scaled RoPE in terms of accuracy and achieves lower latency, indicating the potential of our approach to scale effectively with larger models.
\begin{table}[h]
\setlength{\arrayrulewidth}{1.5pt}
\setlength{\heavyrulewidth}{2pt}
\setlength{\lightrulewidth}{1.5pt}
\centering
\Huge
\resizebox{\columnwidth}{!}{
\begin{tabular}{lcccccc}
\toprule
\multirow{2}{*}{Method} & \multicolumn{2}{c}{En.Sum} & \multicolumn{2}{c}{En.QA} & \multicolumn{2}{c}{R.KV} \\
\cmidrule(lr){2-3} \cmidrule(lr){4-5} \cmidrule(lr){6-7}
 & Acc. (\%) &Time (s) & Acc. (\%) & Time (s) & Acc. (\%) & Time (s) \\
\midrule
\textit{NTK (SPDA)} & 23.49 & 199.52 & 28.77 & 145.69 & 50.00 & 111.98 \\
TokenSelect & \textbf{25.07} & \textbf{114.24} & \textbf{29.91} & \textbf{71.98} & \textbf{88.12} & \textbf{63.27} \\
\bottomrule
\end{tabular}
}
\vspace{-6pt}
\caption{Performance and latency comparison 
on \texttt{Qwen2-72B-Instruct} with tensor parallelism size: 4.}
\vspace{-10pt}
\label{tab: longbench_72b}
\end{table}

\newpage
\section{Formal Statement and Proof of Lemma}
\label{sec:app_lemma1}
\textbf{Lemma 1} (Invariant Top-$k$ Key Selection under Cosine Similarity Threshold, Formal).

\textbf{Assumptions:}
\begin{enumerate}
    \item Let $\mathbf{q}_1, \mathbf{q}_2 \in \mathbb{R}^{d}$ be two query vectors.
    \item Let $\{\mathbf{k}_i\}_{i=1}^{N} \subset \mathbb{R}^{d}$ be a finite set of key vectors.
    \item Let $k$ be a positive integer such that $1 \leq k \leq N$.
    \item Define the cosine similarity between vectors $\mathbf{a}, \mathbf{b} \in \mathbb{R}^{d}$ as:
    \[
    \cos(\mathbf{a}, \mathbf{b}) = \frac{\mathbf{a}\mathbf{b}}{\|\mathbf{a}\|_2\|\mathbf{b}\|_2},
    \]
    where $\|\cdot\|_2$ denotes the Euclidean norm.
    \item Define the top-$k$ selection function based on dot product similarity as: $\mathcal{I}(\mathbf{q}) = \arg\max_{S \subseteq \{1, 2, \dots, N\}, |S|=k} \sum_{i \in S} \mathbf{q} \cdot \mathbf{k}_i$.
    Assume that for any query vectors $\mathbf{q}$, the top-$k$ set $\mathcal{I}(\mathbf{q})$ is uniquely determined.
    \item Let $\epsilon \in (0, 1]$ be a predefined threshold.
\end{enumerate}

\textbf{Lemma Statement:} \textit{If the cosine similarity between the two query vectors $\mathbf{q}_1$ and $\mathbf{q}_2$ satisfies
\[
\cos(\mathbf{q}_1, \mathbf{q}_2) > \epsilon,
\]
then the indices of the top-$k$ keys selected by $\mathbf{q}_1$ and $\mathbf{q}_2$ are identical, i.e.,
\[
\mathcal{I}(\mathbf{q}_1) = \mathcal{I}(\mathbf{q}_2).
\]
}

\textbf{Proof:} We start with the given condition:
\begin{align*}
    \min_{1\le i\le k}{\mathbf{q}_1\mathbf{k}_i} - \max_{j>k}{\mathbf{q}_1\mathbf{k}_j} &> \eta, 
\end{align*}
which we aim to use to demonstrate that:
\begin{align*}
    \min_{1\le i\le k}{\mathbf{q}_2\mathbf{k}_i} - \max_{j>k}{\mathbf{q}_2\mathbf{k}_j} &> 0.
\end{align*}
To facilitate our analysis, we introduce the following notations:
\begin{align*}
    \hat{\eta} = \frac{\eta}{\|\mathbf{q}_1\|},\ \ \ \  
    \hat{\mathbf{q}}_1 = \frac{\mathbf{q}_1}{\|\mathbf{q}_1\|},\ \ \ \ 
    \hat{\mathbf{q}}_2 = \frac{\mathbf{q}_2}{\|\mathbf{q}_2\|}.  
\end{align*}
With these definitions, the original condition becomes:
\begin{align*}
    \min_{1\le i\le k}{\hat{\mathbf{q}}_1\mathbf{k}_i} - \max_{j>k}{\hat{\mathbf{q}}_1\mathbf{k}_j} &> \hat{\eta},
\end{align*}
and our goal transforms to showing:
\begin{align*}
    \min_{1\le i\le k}{\hat{\mathbf{q}}_2\mathbf{k}_i} - \max_{j>k}{\hat{\mathbf{q}}_2\mathbf{k}_j} &> 0.
\end{align*}
Next, let $\theta$ denote the angle between $\mathbf{q}_1$ and $\mathbf{q}_2$, $\cos{\theta} = \hat{\mathbf{q}}_1 \cdot \hat{\mathbf{q}}_2$. We can further define:
\begin{align*}
    \mathbf{p}_1 = \mathbf{q}_2 - \mathbf{q}_1\cos{\theta},\ \ \ \
    \hat{\mathbf{p}}_1 = \frac{\mathbf{p}_1}{\|\mathbf{p}_1\|},
\end{align*}
then $\sin{\theta} = \hat{\mathbf{p}}_1 \cdot \hat{\mathbf{q}}_2$, and
\begin{align*}
    \hat{\mathbf{q}}_2 = \hat{\mathbf{q}}_1 \cos{\theta} + \hat{\mathbf{p}}_1 \sin{\theta}.
\end{align*}
Then we have:
\begin{align*}
    \min_{1\le i\le k}{\hat{\mathbf{q}}_2\mathbf{k}_i} &= \min_{1\le i\le k}{(\hat{\mathbf{q}}_1 \cos{\theta} + \hat{\mathbf{p}}_1 \sin{\theta})\mathbf{k}_i}, \\
    &\ge \min_{1\le i\le k}{\hat{\mathbf{q}}_1 \mathbf{k}_i \cos{\theta}} + \min_{1\le i\le k}{\hat{\mathbf{p}}_1 \mathbf{k}_i \sin{\theta}}, \\
    &\ge \hat{\mathbf{q}}_1 \mathbf{k}_k \cos{\theta} - \|\mathbf{k}\|_{\text{max}}\sin{\theta},
\end{align*}
and 
\begin{align*}
    \max_{j>k}{\hat{\mathbf{q}}_2\mathbf{k}_j} &= \max_{j>k}{(\hat{\mathbf{q}}_1 \cos{\theta} + \hat{\mathbf{p}}_1 \sin{\theta})\mathbf{k}_j} \\
    &\le \max_{j>k}{\hat{\mathbf{q}}_1 \mathbf{k}_i \cos{\theta}} + \max_{j>k}{\hat{\mathbf{p}}_1 \mathbf{k}_i \sin{\theta}}, \\
    &\le \hat{\mathbf{q}}_1 \mathbf{k}_{p+1} \cos{\theta} + \|\mathbf{k}\|_{\text{max}}\sin{\theta}.
\end{align*}
Therefore, 
\begin{align}
    \min_{1\le i\le k}{\hat{\mathbf{q}}_2\mathbf{k}_i} &- \max_{j>k}{\hat{\mathbf{q}}_2\mathbf{k}_j} \nonumber \\ 
    &\ge \hat{\mathbf{q}}_1 \mathbf{k}_p \cos{\theta} - \|\mathbf{k}\|_{\text{max}}\sin{\theta} \nonumber \\ 
    & \quad \quad \quad \quad \quad \;\; -(\hat{\mathbf{q}}_1 \mathbf{k}_{p+1} \cos{\theta} \nonumber \\ 
    & \quad \quad \quad \quad \quad \;\; + \|\mathbf{k}\|_{\text{max}}\sin{\theta}) \nonumber \\
    &= (\hat{\mathbf{q}}_1 \mathbf{k}_p \cos{\theta}-\hat{\mathbf{q}}_1 \mathbf{k}_{p+1} \cos{\theta}) \nonumber \\ 
    & \quad \quad \quad \quad \quad \quad - 2\|\mathbf{k}\|_{\text{max}}\sin{\theta} \nonumber \\
    & \ge \hat{\eta}\cos{\theta} - 2\|\mathbf{k}\|_{\text{max}}\sin{\theta}. \label{eqn:lemma1}
\end{align}
In order to have Eqn. (\ref{eqn:lemma1}) $> 0$, we require 
\begin{align*}
    &\hat{\eta}\cos{\theta} > 2\|\mathbf{k}\|_{\text{max}}\sin{\theta}, \\
    \Rightarrow \ &\frac{\sin{\theta}}{\cos{\theta}} < \frac{\hat{\eta}}{2\|\mathbf{k}\|_{\text{max}}}, \\
    \Rightarrow \ &\frac{1-\cos^2{\theta}}{\cos^2{\theta}} < \Big{(}\frac{\hat{\eta}}{2\|\mathbf{k}\|_{\text{max}}}\Big{)}^2, \\
    \Rightarrow \ &\cos{\theta} \ge \frac{1}{\sqrt{{1+\Big(\frac{\hat{\eta}}{2\|\mathbf{k}\|_{\text{max}}}\Big)^2}}}.
\end{align*}
This final inequality establishes a sufficient condition for the original statement to hold, thereby completing the proof.

\section{Overview of LLMs Inference}
\label{sec:llm_inference}
Nowadays, mainstream LLMs are primarily based on the Decoder-only Transformer architecture. Each transformer layer includes a multi-head attention (MHA) and a feed-forward networks (FFN). The inference process of LLMs can be divided into two stages: the prefill stage and the decode stage. 

The prefill stage is the preparatory phase of the inference process. In this stage, the user's input is processed layer by layer through a single forward pass of LLMs, generating KV Cache for each layer. The generation of KV Cache is completed by the MHA module. Assuming $\mathbf{X}_\text{prefill} \in \mathbb{R}^{n_\text{in} \times d}$ is the input of a transformer layer, where $n_\text{in}$ is the number of tokens in user's input sequence and $d$ is the hidden size. The MHA computation in the prefill stage is as follows (simplified to single head):

{\small
\vspace{-16pt}
\begin{equation}
[\mathbf{Q}_\text{prefill}, \mathbf{K}_\text{prefill}, \mathbf{V}_\text{prefill}] = \mathbf{X}_\text{prefill} \cdot \left[\mathbf{W}_q, \mathbf{W}_k, \mathbf{W}_v\right],
\label{eq:qkv_prefill}
\end{equation}
}

{\small
\vspace{-24pt}
\begin{equation}
\mathbf{O}_\text{prefill} = \text{softmax}\left(\frac{\mathbf{Q}_\text{prefill} \cdot {\mathbf{K}_\text{prefill}}^\top}{\sqrt{d}}\right) \cdot \mathbf{V}_\text{prefill},
\vspace{-6pt}
\label{eq:prefill_o}
\end{equation}
}where $\mathbf{W}_q, \mathbf{W}_k, \mathbf{W}_v$ are linear projections, $[\cdot]$ represents tensor concatenation operation, and \textit{Eq.}(\ref{eq:prefill_o}) is also known as Scaled Dot Product Attention (SDPA). After these computation, $\mathbf{K}_\text{prefill}$ and $\mathbf{V}_\text{prefill}$ are stored as the KV Cache for current layer $\mathbf{K}_\text{cache}$ and $\mathbf{V}_\text{cache}$, and $\mathbf{O}_\text{prefill}$ is used for subsequent calculations.

The decode stage is the phase where LLMs actually generate the response. In the decode stage, LLMs load the KV Cache and generate $n_\text{out}$ output tokens autoregressively through $n_\text{out}$ forward passes. Assuming $\mathbf{X}_\text{decode} \in \mathbb{R}^{1 \times d}$ is the input of a transformer layer in a forward pass, the computation of MHA in the decode stage is as follows (The calculation of $\mathbf{Q}_\text{prefill}$ and $\mathbf{O}_\text{prefill}$ is consistent with that in the prefill stage):

{\small
\vspace{-14pt}
\begin{equation}
\begin{aligned}
\mathbf{K}_\text{decode} &= \left[\mathbf{K}_\text{cache},\ \mathbf{X}_\text{decode} \cdot \mathbf{W}_k\right],\;
\mathbf{K}_\text{cache} \leftarrow \mathbf{K}_\text{decode}, \\
\mathbf{V}_\text{decode} &= \left[\mathbf{V}_\text{cache}, \ \mathbf{X}_\text{decode} \cdot \mathbf{W}_v\right],\;
\mathbf{V}_\text{cache} \leftarrow \mathbf{V}_\text{decode},
\end{aligned}
\vspace{-6pt}
\end{equation}
}where $\mathbf{K}_\text{decode}, \mathbf{V}_\text{decode}$ are composed of the KV Cache and the KV corresponding to the current input, which are then used to update the KV Cache of the current layer for use in the next forward pass.

LLMs inference, unlike training, is memory-bound, necessitating frequent GPU I/O operations between HBM and SRAM while underutilizing processing units. This bottleneck is particularly evident in SDPA computation. Optimizing for I/O is crucial for enhancing LLMs inference efficiency, especially in long-context scenarios.

\begin{table*}[t]
\centering
\small
\resizebox{0.8\textwidth}{!}{
\begin{tabular}{lccccccccc}
\toprule
Method & En.Sum & En.QA & En.MC & Math.F & R.PK & R.Num & R.KV & Avg. \\
\midrule
H2O & 2.8 & 0.7 & 0.0 & 6.0 & 2.5 & 2.4 & 0.0 & 2.1 \\
InfLLM & 24.3 & 19.5 & 43.7 & 23.7 & \textbf{100.0} & \textbf{99.0} & 5.0 & 45.0 \\
\rowcolor{cyan!20} TokenSelect & \textbf{26.9} & \textbf{21.3} & \textbf{45.8} & \textbf{28.2} & \textbf{100.0} & 97.2 & \textbf{48.4} & \textbf{52.5} \\
\bottomrule
\end{tabular}
}
\caption{Performance comparison with H$_2$O \cite{h2o} on \texttt{Llama-3-8B-Instruct}, baseline performance is referenced from \citet{infllm}.}
\label{tab:results-comparison}
\end{table*}
\section{Comparison with Token Eviction-based Methods  (\textit{e.g.}, H$_2$O)}
Token eviction–based methods~\cite{h2o,tova}, led by H$_2$O~\cite{h2o}, have pioneered the field of long-context inference, achieving early state-of-the-art performance. Although both our method and H$_2$O employ token‐level criticality estimation, they fall under two entirely different taxonomies. As discussed in \textit{Sec.}~\ref{sec:related_works} and \textit{Sec.}~\ref{sec:preliminaries}, H$_2$O is a query‐independent KV cache selection method, which suffers from three main drawbacks:

\begin{enumerate}[leftmargin=*, labelsep=4mm]
  \item \emph{Lack of dynamism:} Its importance scoring relies on attention scores from previous queries and keys. Consequently, KV pairs that are crucial for the current query may have been discarded earlier—a phenomenon also confirmed by QUEST~\cite{quest}. \textit{Fig.}~1 and~2 of QUEST provide an intuitive illustration of the differences between query‐based methods (e.g., our \textit{TokenSelect}) and H$_2$O. Notably, \textit{TokenSelect} leverages a dynamic selection strategy, enabling state‐of‐the‐art performance with a minimal token budget.
  
  \item \emph{Inability to extend sequence length:} Since H$_2$O depends on the model’s original attention mechanism, it cannot extend the effective context length. In contrast, our approach can easily extend a model with an original maximum length of 4K–32K tokens to an effective length exceeding 1M tokens.
  
  \item \emph{Inefficient implementation:} H$_2$O evaluates token importance based on attention scores, making it incompatible with efficient kernels such as FlashAttention~\cite{flashattention2}. This limitation restricts its scalability. Our method, however, is designed for broad compatibility and is fully transparent to large‐scale inference acceleration infrastructures, including paged attention, tensor parallelism, and prefix caching, making it ready for large‐scale online serving.
\end{enumerate}

To further demonstrate the superiority of \textit{TokenSelect}, we present experimental results in Table~\ref{tab:results-comparison}. These results corroborate the findings of previous studies \cite{quest,infllm}, showing that query‐independent methods are inferior to query‐based approaches.

\section{Detailed Descriptions on Baselines}
\label{sec:des_baseline}
In this paper, we use the following baselines:
\begin{itemize}[leftmargin=*, labelsep=4mm]
    \item \textbf{NTK-Aware Scaled RoPE}~\cite{ntk-rope}: A nonlinear RoPE interpolation method.
    \item \textbf{SelfExtend}: A RoPE interpolation method that reuses the position ids of neighboring tokens.
    \item \textbf{StreamingLLM}~\cite{streamllm}: The state-of-the-art method for long-context inference with predefined sparse patterns. Similar approaches include \textbf{LM-Infinite}~\cite{lm-inf}.
    \item \textbf{InfLLM}~\cite{infllm}: The state-of-the-art method for long-context inference and length extrapolation using a block-level selective sparse attention method.
    \item \textbf{MInference}~\cite{minference}: The state-of-the-art method for long-context prefilling acceleration, utilizing three sparse patterns including block-level sparse attention.
    \item \textbf{SnapKV}~\citep{snapkv}: A fine-tuning-free approach that efficiently compresses KV caches by selecting clustered important KV positions for each attention head.
    \item \textbf{InfiniGen}~\cite{infinigen}: A KV cache management framework that reduces memory overhead in offloading-based LLM inference by prefetching only essential KV cache entries through selective token rehearsal.
    \item \textbf{QUEST}~\cite{quest}: A query-aware KV cache management algorithm by selecting critical KV cache based on the query-aware sparsity at page granularity.
    \item \textbf{RetrievalAttention}~\cite{retrievalattention}: The state-of-the-art method leveraging approximate nearest neighbor search on CPU memory and an attention-aware vector search algorithm to address distribution mismatches.
\end{itemize}

\section{More Information on Datasets}
\label{sec:app_metrics}
In this paper, we use the following datasets: \begin{itemize}[leftmargin=*, labelsep=4mm]
\item \textbf{InfiniteBench}~\cite{infbench}: The mainstream long-context benchmark consisting of multi-tasks. The average length of it exceeds 200K tokens.
\item \textbf{RULER}~\cite{ruler}: A challenging long-context benchmark containing 13 different tasks, with subsets of varying lengths up to 128K tokens.
\item \textbf{LongBench}~\cite{longbench}: Another mainstream long-context benchmark comprising 6 types of tasks. The 95\% percentile for its lengths is 31K tokens.
\end{itemize} 
For InfiniteBench~\cite{infbench}, we use longbook\_sum\_eng (En.Sum), longbook\_qa\_eng (En.QA), longbook\_choice\_eng
(En.MC), longdialogue\_qa\_eng (En.Dia), code\_debug (Code.D), math\_find (Math.F), passkey (R.PK), number\_string (R.Num) and kv\_retrieval (R.KV) as evaluation datasets. The corresponding evaluation metrics are shown in Table~\ref{tab: infinitebench_metrics}. 
RULER~\cite{ruler} consists of various evaluation tasks:  Single NIAH (needle in a haystack), Multi-keys NIAH, Multi-values NIAH, Multi-values NIAH, Multi-queries NIAH, Variable 
Tracking, Common Words Extraction, Frequent Words Extraction and 
Question
Answering. The evaluation metric is match rate. For LongBench, we use all English tasks with evaluation metrics in Table~\ref{tab: longbench_metrics}.

\section{Comparison on Prefill Latency}
We note that MInference~\cite{minference} has gained widespread adoption in real-world long-context inference applications due to its novel design of attention sparse patterns and efficient implementation based on vLLM. In the main text, we demonstrated \textit{TokenSelect}'s performance advantages. To further prove its efficiency readiness for real-world applications, we followed Minference's approach by comparing the end-to-end prefill latency under paged KV Cache management for different input token lengths on \texttt{Llama-3-8B} using a single A100, with results shown in Table~\ref{tab:prefill_time}. The results indicate that \textit{TokenSelect} demonstrates significant advantages with shorter input token lengths, while maintaining efficiency comparable to MInference as input token lengths increase.
\begin{table}[h]
\centering
\small
\resizebox{\columnwidth}{!}{%
\begin{tabular}{lccc}
\toprule
\multirow{2}{*}{Length} & \multirow{2}{*}{\begin{tabular}{c}FlashAttention-2\\(vLLM)\end{tabular}} & \multirow{2}{*}{\begin{tabular}{c}MInference\\(vLLM)\end{tabular}} & \multirow{2}{*}{TokenSelect} \\
                        &                                                                       &                                                                       &                   \\
\midrule
1K    & 0.081   & 3.017   & 0.092 \\
10K   & 0.832   & 2.762   & 1.290 \\
50K   & 7.717   & 7.540   & 5.712 \\
100K  & 21.731  & 14.081  & 12.088 \\
128K  & 32.863  & 18.827  & 15.920 \\
200K  & OOM     & OOM     & 26.500 \\
300K  & OOM     & OOM     & 43.406 \\
\bottomrule
\end{tabular}%
}
\caption{Comparison of end-to-end prefill latency (s).}
\label{tab:prefill_time}
\end{table}

\section{Detailed Performance Comparisons Under Different Cache Threshold $\theta$}
\label{sec:detail_theta}
Table~\ref{tab: performance_theta} presents the performance sensitivity to the threshold $\theta$ of the Selection Cache across various tasks. The results indicate that although $\theta$-sensitivity varies across different task types, most tasks exhibit only slight performance degradation as $\theta$ decreases. This suggests potential for further accelerating \textit{TokenSelect}’s decode stage by reducing $\theta$ in the vast majority of cases. It is worth noting, however, that more challenging retrieval tasks—such as R.KV—show noticeable performance degradation as $\theta$ decreases, indicating higher dynamicity requirements for token selection in these tasks.
\begin{table*}[t]
\centering
\tiny
\resizebox{1.0\textwidth}{!}{
\begin{tabular}{lcccccccccc}
\toprule
$\theta$ & En.Sum & En.QA & En.MC & En.Dia & Code.D & Math.F & R.PK & R.Num & R.KV & Avg. \\
\midrule
0.5  & 20.99 & 17.83 & 54.31 & 7.50 & 30.20 & 21.14 & 100.00 & 96.10 & 0.20  & 38.69 \\
0.6  & 21.21 & 18.08 & 54.31 & 7.50 & 30.20 & 21.36 & 100.00 & 96.78 & 0.20  & 38.84 \\
0.7  & 20.73 & 18.08 & 54.31 & 7.50 & 30.46 & 21.36 & 100.00 & 98.98 & 4.40  & 39.53 \\
0.8  & 21.47 & 17.85 & 54.31 & 7.50 & 30.20 & 21.58 & 100.00 & 100.00 & 24.60 & 41.94 \\
0.85 & 22.39 & 18.15 & 54.31 & 7.50 & 30.20 & 21.79 & 100.00 & 100.00 & 68.20 & 46.94 \\
0.9  & 22.62 & 18.86 & 54.31 & 7.50 & 30.20 & 21.71 & 100.00 & 100.00 & 86.60 & 49.08 \\
0.95 & 22.46 & 18.54 & 54.31 & 7.50 & 30.56 & 21.77 & 100.00 & 100.00 & 86.20 & 49.05 \\
1.0  & 22.66 & 18.68 & 54.31 & 7.50 & 30.51 & 21.78 & 100.00 & 100.00 & 86.84 & 49.15 \\
\bottomrule
\end{tabular}
}
\vspace{-2pt}
\caption{Performance using different selection cache similarity thresholds using \texttt{Qwen2-7B-Instruct}.}
\label{tab: performance_theta}
\end{table*}

\section{Experimental Results on LongBench}
\label{sec:app_longbench}
Compared to InfiniteBench and RULER, LongBench has much shorter text lengths. The 95\% percentile for its lengths is 31K tokens. Considering that recent LLMs after SFT generally have context lengths of up to 32K tokens~\cite{qwen}, LongBench is less suitable for evaluating state-of-the-art long-context inference methods. Nevertheless, as shown in Table \ref{tab: longbench}, our \textit{TokenSelect} still demonstrates superior overall performance compared to most baseline methods. It's worth noting that \texttt{Yi-1.5-6B} did not yield effective results on the SAMSum task because it failed to correctly follow instructions.

\section{Use of AI Assistants}
In this paper, AI Assistants were used for literature retrieval and grammar checking.

\begin{table*}[t]
\small
\centering
\resizebox{1.0\textwidth}{!}{
    \begin{tabular}{l|ccccccccc}
    \toprule
    Datasets & En.Sum & En.QA & En.MC & En.Dia & Code.D & Math.F & R.PK & R.Num & R.KV \\
    Metrics & Rouge-L-Sum & QA F1 Score & Accuracy & Accuracy & Accuracy & Accuracy & Accuracy & Accuracy & Accuracy\\
    \bottomrule
    \end{tabular}
}
\caption{Evaluation metrics of different datasets on InfiniteBench.}
\label{tab: infinitebench_metrics}
\end{table*}

\begin{table*}[t]
\small
\centering
\resizebox{1.0\textwidth}{!}{
    \begin{tabular}{l|cccccccc}
    \toprule
    Datasets & NQA & Qasper &  MFQA & HQA & 2WikiMQA & Musique & GovReport & QMSum \\
    Metrics & QA F1 Score & QA F1 Score & QA F1 Score & QA F1 Score & QA F1 Score & QA F1 Score & Rouge-L & Rouge-L \\
    \midrule
    Datasets & MultiNews 
    & TREC & TQA & SAMSum & PsgCount & PsgRetrieval & LCC & RepoBench-P \\
    Metrics & Rouge-L & Accuracy & QA F1 Score & Rouge-L & Accuracy & Accuracy & Code Sim Score & Code Sim Score \\
    \bottomrule
    \end{tabular}
}
\caption{Evaluation metrics of different datasets on LongBench.}
\label{tab: longbench_metrics}
\end{table*}

\begin{table*}[t]
\small
\centering
\resizebox{1.0\textwidth}{!}{
    \begin{tabular}{lcccccccccc}
    \toprule
    \rowcolor{gray!20} Methods & NQA & Qasper &  MFQA & HQA & 2WikiMQA & Musique & GovReport & QMSum & MultiNews & \\
    \midrule
    \textit{Qwen2-7B} & 24.24 & 45.42 & 47.79 & 42.76 & 44.38 & 24.16 & 33.80 & 23.78 & 26.17\\
    NTK & 26.25 & 45.94 & 50.76 & 53.20 & 50.31 & 30.83 & 32.75 & 23.21 & 25.94 \\
    SelfExtend & 7.15 & 20.37 & 24.06 & 14.91 & 13.73 & 4.75 & 16.92 & 16.53 & 18.74\\
    StreamLLM & 19.49 & 42.56 & 39.63 & 42.43 & 44.67 & 15.22 & 31.51 & 20.57 & 26.00 \\
    InfLLM & 27.47 & 41.44 & 46.99 & 47.47 & 49.29 & 25.62 & 32.68 & 23.10 & 26.77\\
    \rowcolor{cyan!20} TokenSelect & 24.18 & 42.29 & 45.77 & 48.62 & 49.08 & 27.85 & 33.69 & 23.03 & 26.35\\
    \midrule
    \textit{Llama-3-8B} & 19.85 & 42.36 & 41.03 & 47.38 & 39.20 & 22.96 & 29.94 & 21.45 & 27.51 \\
    NTK & 9.90 & 45.35 & 49.41 & 48.86 & 29.22 & 24.56 & 34.31 & 23.82 & 27.27 \\
    SelfExtend & 1.72 & 8.90 & 20.80 & 8.65 & 6.97 & 3.27 & 13.99 & 15.36 & 17.66 \\
    StreamLLM & 20.05 & 42.46 & 39.54 & 43.69 & 37.89 & 19.68 & 29.17 & 21.33 & 27.56 \\
    InfLLM & 22.64 & 43.70 & 49.03 & 49.04 & 35.61 & 26.06 & 30.76 & 22.70 & 27.57 \\
    \rowcolor{cyan!20} TokenSelect & 22.44 & 40.74 & 47.73 & 50.33 & 31.38 & 24.53 & 32.56 & 23.50 & 27.92 \\
    \midrule
    \textit{Yi-1.5-6B} & 17.18 & 32.56 & 39.06 & 36.26 & 39.25 & 16.32 & 30.53 & 20.21 & 26.20 \\
    NTK & 0.80 & 35.06 & 29.05 & 7.47 & 24.38 & 0.73 & 13.66 & 6.25 & 25.43 \\
    SelfExtend & 3.29 & 19.03 & 26.00 & 17.11 & 11.88 & 7.73 & 20.38 & 17.46 & 21.79 \\
    StreamLLM & 15.05 & 33.27 & 38.31 & 34.91 & 36.92 & 16.33 & 29.38 & 20.02 & 26.14 \\
    InfLLM & 17.65 & 36.25 & 45.40 & 41.25 & 35.89 & 16.94 & 30.22 & 20.85 & 26.04 \\
    \rowcolor{cyan!20} TokenSelect & 19.36 & 33.98 & 48.14 & 45.05 & 40.13 & 22.98 & 31.59 & 21.51 & 26.48 \\
    \midrule
    \rowcolor{gray!20} Methods & TREC & TQA & SAMSum & PsgCount & PsgRetrieval & LCC & RepoBench-P & \multicolumn{2}{@{}c}{\bf Average} \\
    \midrule
    \textit{Qwen2-7B} & 78.50 & 88.77 & 46.33 & 5.50 & 70.00 & 62.40 & 61.95 & \multicolumn{2}{@{}c}{45.37}\\
    NTK & 79.50 & 89.51 & 46.03 & 5.50 & 60.00 & 59.36 & 59.69 & \multicolumn{2}{@{}c}{46.17}\\
    SelfExtend & 16.50 & 27.54 & 29.42 & 4.50 & 0.00 & 41.42 & 41.89 & \multicolumn{2}{@{}c}{18.65} \\
    StreamLLM & 75.50 & 87.19 & 46.27 & 3.50 & 27.50 & 61.18 & 61.12 & \multicolumn{2}{@{}c}{40.27}\\
    InfLLM & 70.50 & 87.51 & 44.53 & 4.00 & 46.50 & 55.08 & 57.53 & \multicolumn{2}{@{}c}{42.90}\\
    \rowcolor{cyan!20} TokenSelect & 74.00 & 89.26 & 45.94 & 5.00 & 42.50 & 61.48 & 59.33 & \multicolumn{2}{@{}c}{43.64}\\
    \midrule
    \textit{Llama-3-8B} & 74.00 & 90.50 & 42.30 & 8.50 & 62.50 & 60.83 & 49.14 & \multicolumn{2}{@{}c}{42.46} \\
    NTK & 73.00 & 88.74 & 42.51 & 8.87 & 99.50 & 33.62 & 35.04 & \multicolumn{2}{@{}c}{42.12} \\
    SelfExtend & 20.50 & 16.82 & 25.39 & 5.75 & 7.50 & 26.24 & 31.22 & \multicolumn{2}{@{}c}{14.42} \\
    StreamLLM & 73.50 & 90.08 & 41.55 & 5.00 & 49.00 & 60.35 & 48.95 & \multicolumn{2}{@{}c}{40.61}\\
    InfLLM & 73.50 & 90.91 & 42.43 & 7.17 & 84.00 & 59.88 & 46.48 & \multicolumn{2}{@{}c}{44.46}\\
    \rowcolor{cyan!20} TokenSelect & 67.50 & 92.22 & 42.16 & 4.54 & 87.00 & 58.86 & 51.24 & \multicolumn{2}{@{}c}{44.04} \\
    
   \midrule
   \textit{Yi-1.5-6B} & 71.50 & 48.79 & 0.79 & 3.00 & 28.50 & 57.10 & 52.53 & \multicolumn{2}{@{}c}{32.48} \\
    NTK & 40.00 & 12.71 & 1.34 & 0.50 & 3.35 & 54.55 & 37.24 & \multicolumn{2}{@{}c}{18.28} \\
    SelfExtend & 23.75 & 30.61 & 2.58 & 2.75 & 13.50 & 43.17 & 35.45 & \multicolumn{2}{@{}c}{18.53} \\
    StreamLLM & 69.00 & 73.36 & 0.82 & 2.50 & 18.50 & 56.37 & 49.05 & \multicolumn{2}{@{}c}{32.49} \\
    InfLLM & 71.50 & 71.49 & 1.01 & 4.00 & 10.50 & 56.88 & 46.28 & \multicolumn{2}{@{}c}{33.25} \\
    \rowcolor{cyan!20} TokenSelect & 62.50 & 69.70 & 0.62 & 3.50 & 41.50 & 54.32 & 54.99 & \multicolumn{2}{@{}c}{36.02} \\
    
    \bottomrule
    \end{tabular}
}
\caption{Comparison of different methods with different origin models on LongBench.}
\label{tab: longbench}
\end{table*}

\end{document}